\documentclass[conference]{IEEEtran}

\usepackage{cite}

\usepackage{graphicx}
\usepackage{subfig} 
\graphicspath{{images/}}
\usepackage{amsmath}
\usepackage{amssymb}
\usepackage{amsthm}
\newtheorem{theorem}{Theorem}

\providecommand{\abs}[1]{|#1|} 
\newenvironment{proofsketch}{\trivlist\item[]\emph{Proof Sketch}:}%

\begin{document} 
 
 \title{Bayesian Discovery of Multiple Bayesian Networks\\ via Transfer Learning} 

\author{\IEEEauthorblockN{Diane Oyen}
\IEEEauthorblockA{University of New Mexico}
\IEEEauthorblockA{doyen@cs.unm.edu}
\and
\IEEEauthorblockN{Terran Lane}
\IEEEauthorblockA{Google, Inc}
\IEEEauthorblockA{terran.lane@gmail.com}
}


%


\maketitle 
 
\begin{abstract} 
Bayesian network structure learning algorithms with limited data are being used in domains such as systems biology and neuroscience to gain insight into the underlying processes that produce observed data. Learning reliable networks from limited data is difficult, therefore transfer learning can improve the robustness of learned networks by leveraging data from related \emph{tasks}. Existing transfer learning algorithms for Bayesian network structure learning give a single maximum a posteriori estimate of network models. Yet, many other models may be equally likely, and so a more informative result is provided by Bayesian structure discovery. Bayesian structure discovery algorithms estimate posterior probabilities of structural features, such as edges. We present transfer learning for Bayesian structure discovery which allows us to explore the shared and unique structural features among related tasks. Efficient computation requires that our transfer learning objective factors into local calculations, which we prove is given by a broad class of transfer biases. Theoretically, we show the efficiency of our approach. Empirically, we show that compared to single task learning, transfer learning is better able to positively identify true edges. We apply the method to whole-brain neuroimaging data.
\end{abstract} 
 
\section{Introduction}

The discovery of structural features in Bayesian networks is of great interest in scientific domains such as bioinformatics and neuroscience. The goal is to understand the relationships among variables in a system, such as genes in a gene expression network or activity levels of regions of the brain in functional brain networks. However, the data collected is often done in several separate but related experiments. Therefore, the full data set is actually composed of several distinct, but related, subsets of data --- called \emph{tasks} in transfer learning. For each task there may not be enough samples to learn a robust model. Transfer learning leverages information among tasks to smooth learned models \cite{thrun_is_1996,caruana_multitask_1997}. These smoothed models tend to be more robust to sample noise and generalize to holdout data better than models learned without transfer. Furthermore, in unsupervised learning, the set of models learned among tasks will share many features in common, easing interpretation of the models. Differences among the learned task-specific models are more likely to be due to real differences in the generating distribution of the data rather than spurious differences \cite{niculescu-mizil_inductive_2007,zhang_multitask_2010}.

Existing transfer Bayesian network learning algorithms have two major limitations: 1) they use \emph{heuristic} search over the space of sets of graphs; and 2) produce a \emph{single} point maximum a posteriori model. Yet, there may be many other solutions of similar likelihood and therefore a point solution will not give a full picture of likely relationships among variables. A point solution can perform well at predicting future data, but it is misleading to present such a solution to the domain expert as the only model that explains the data. Instead, we would like to learn a posterior distribution over solutions and extract meaningful summary statistics about features of interest. Such algorithms exist for learning individual Bayesian networks and are referred to as \emph{network discovery} \cite{friedman_being_2003,Koivisto:2004:EBS:1005332.1005352}. Extending these algorithms for transfer network learning is not trivial, as we explain in the next paragraph.

Algorithms for estimating the posterior probability distribution for a single network generally fall into two categories: those that search over \emph{structure} space and those that search over \emph{order} space. Structure-space algorithms make small local changes to the learned graph structure (typically, the addition, removal or reversal of a single edge). For these small changes, updating the likelihood of the graph, and therefore the posterior distribution, is fast but covering the full space of structures can be slow and can get stuck in local maxima \cite{madigan1995bayesian,grzegorczyk_improving_2008}. Extending those structure-search algorithms to multiple tasks would explode the model search space exponentially, exacerbating convergence issues. On the other hand, order-search algorithms exploit the tractability of calculating posteriors given a fixed ordering of the variables (described in more detail later). Node order dictates which nodes are allowed to be parents to any given node, and therefore changes in node order are more global than structure-space changes. There are relatively efficient algorithms for calculating exact posteriors of structural features \cite{Koivisto:2004:EBS:1005332.1005352,Parviainen:2009:ESD:1795114.1795165} or approximate posteriors \cite{friedman_being_2003,Niinimaki11} that have been shown to be faster than structure-search. However, there is no \emph{structural} prior provided in these order-space formulations; instead there is a prior over orders. We do not want to impose transfer at the level of \emph{orders}, but rather at the level of structures (a particular edge appearing in one task will be preferred in other tasks).

Our main challenge, therefore, is to incorporate a structural bias term into the order-search formulation. With such a bias term, we can impose a transfer bias to learn more robust networks, while leveraging the most efficient Bayesian discovery algorithms that currently exist. Our major contribution is proving that structural bias can be efficiently incorporated into the order-conditioned network discovery formulation. We prove that our transfer formulation factors into local calculations. Thus, we provide the first transfer algorithm that can calculate \emph{exact} posteriors for multiple networks of moderate size. We are also the first to show how transfer can incorporated into state-of-the-art order-search approximation algorithms for larger networks.


Our contribution is a proof that structure bias can be efficiently incorporated into order-conditioned Bayesian structure discovery: a necessary requirement for using the efficient algorithms of network discovery \cite{Koivisto:2004:EBS:1005332.1005352, friedman_being_2003}. This is a finding that can impact many structure discovery problems. We give a specific formulation of \emph{multitask} Bayesian network discovery that uses the structure bias to transfer information among tasks. We further show that we can take a Bayesian approach to average over all possible settings of the transfer parameter rather than needing to select this parameter. Empirical results on networks, of size 8 variables and 37 variables, indicate that our transfer approach learns posterior probabilities that are closer to the optimal values than single-task learning algorithms. We apply our multitask algorithm to neuroimaging data with 150 variables and demonstrate that the multitask algorithm produces fewer spurious edges than the non-multitask algorithm while providing more knowledge discovery information than the standard point solution.

\section{Related Work}

Network structure discovery in the face of limited data is an extensively studied problem. With limited data, the posterior probability of even the optimal network may be quite small, however Friedman and Koller (2003) show that the marginal posterior probabilities over subgraphs or structural features can be quite high given the same data \cite{friedman_being_2003}. They propose the so-called order-MCMC algorithm for estimating such posterior probabilities. Koivisto and Sood (2004)  give a dynamic programming method for calculating exact posterior probabilities of network features conditioned on orders \cite{Koivisto:2004:EBS:1005332.1005352}. Further improvements are made to make the approach more memory efficient \cite{Parviainen:2009:ESD:1795114.1795165} and to produce partial-order MCMC  \cite{Niinimaki11}. We show how to extend these single-task learning approaches to the transfer learning problem.


Transfer and multitask learning leverage information among related problems called \emph{tasks} \cite{thrun_is_1996,caruana_multitask_1997}. Formulations for multitask learning of Bayesian networks exist \cite {niculescu-mizil_inductive_2007,oyen_aaai_2012}. Starting from these, we derive an inductive bias toward similar structures among related tasks. However, the solutions to the multitask problems in these papers are found through heuristic search, producing a point estimate rather than a posterior distribution, which we desire. They also require a parameter that determines the strength of transfer bias. 

Another approach to learning Bayesian networks from limited data uses the concept of network discovery with a prior over structures obtained from domain knowledge \cite{grzegorczyk_improving_2008,werhli2007reconstructing}. However, these approaches resort to using MCMC in structure space (rather than order space) to avoid the difficulties in assigning priors conditioned on orders. Furthermore, they do not solve the problem considered here of multiask learning.

\section{Preliminaries}

First, we introduce background information about Bayesian structure discovery for learning a single task and then describe MAP multitask Bayesian network objectives. We combine ideas from both of these approaches to produce Bayesian structure discovery of multitask Bayesian networks.

Bayesian networks compactly describe joint probability distributions by encoding conditional independencies in multivariate data. A Bayesian network $B = \{G, \theta\}$ describes the joint probability distribution over $n$ random variables $\mathbf{X} = [X_1, X_2, \ldots, X_n]$, where $G$ is a directed acyclic graph (DAG) and the conditional probability distributions are parameterized by $\theta$ \cite{heckerman_learning_1995}. An edge $(X_i,X_j)$ in $G$ means that the child $X_j$ is conditionally independent of all non-descendants given its parent $X_i$. The \emph{structure} of the network, $G $, is of particular interest in many domains as it is easy to interpret and gives valuable information about the interaction of variables.

\subsection{Structural Feature Discovery}

Given a limited amount of data, the posterior probability of any network may be quite small. However, summary statistics regarding structural features of networks may have high posterior even with limited data \cite{friedman_being_2003}. Structural features (such as an edge) can be described by an indicator function $f$ such that for $f(G)=1$ the feature exists in graph $G$, otherwise $f(G)=0$. The posterior probability of the feature is equivalent to the expectation of its indicator, $P(f|D) = \sum_G P(G|D) f(G)$. However, this sum can be intractable, as the number of DAGs is super-exponential in the number of variables.

An important insight to making this sum tractable is that we could fix the \emph{order} of the variables. An order, $\prec$, is a permutation on the indices of the variables $X_{\prec(1)}, X_{\prec(2)}, \ldots, X_{\prec(n)}$ such that parents must precede children in the order, i.e. $X_j$ cannot be a parent of $X_i$ if $\prec \!\! (j) \geq\; \prec \!\! (i)$. Given an order, learning optimal parents for each child factors into local calculations, and summing over DAGs consistent with the order is tractable \cite{buntine1991theory,cooper_bayesian_1992}. \cite{friedman_being_2003} condition on a node order, and then obtain the unconditional posterior by summing over orders:
\begin{equation*}
P(f|D) = \frac{1}{P(D)}\sum_\prec P(\! \prec)\! \sum_{G \subseteq \! \prec}\! P(D|G) P(G | \! \prec) f(G)
\end{equation*}
Note that these two formulations for $P(f|D)$ are not the same, as most DAGs, $G$, will be consistent with multiple orders, $\! \prec$. Typically, this formulation produces an acceptable bias in favor of simpler structures.

\cite{Koivisto:2004:EBS:1005332.1005352} give an efficient method for calculating this sum. The approach is rather involved, so we summarize only the key points here. They make several reasonable assumptions, then break the calculation into three steps. First, we describe the modularity assumptions:
\begin{enumerate}
\item \textbf{Parameter modularity}: Modularity of the Bayesian network parameters must also hold, $P(\theta | G) = \prod_{i=1}^n P(\theta_{i,\pi_i} | \pi_i)$ and $P(X = x | G) = \prod_{i=1}^n P(x_i | x_{\pi_i}, \theta_{i, \pi_i})$.

\item \textbf{Structure prior modularity}: The network model prior must be modular so that $P(G, \! \prec) = c \prod_{i=1}^n P(U_i) P(\pi(X_i))$, where $U_i$ is the set of variables preceding $X_i$ in the order $\! \prec$ (potential parents of $X_i$) and $c$ is a normalization constant.

\item \textbf{Feature modularity}: The features must be modular, $f(G) = \prod_{i=1}^n f_i(\pi(X_i))$ where $\pi(X_i)$ is the parent set of variable $i$.
\end{enumerate}



The most common feature to look for is a directed edge $u \rightarrow v$ s.t. $f = 1 \text{ if } X_u \in \pi_v$, which is clearly modular. If these modularity assumptions hold, then the likelihood over order space factors into local calculations as shown in Eq~\ref{eq:modularSTL} \cite{Koivisto:2004:EBS:1005332.1005352}.
\begin{equation}
\begin{split}
P(f, D | \! \prec) &= \prod_{i=1}^n \sum_{\pi_i \subseteq U_i}
	P(\pi_i | U_i) P(x_i | \pi_i) f_i(\pi_i)\\
P(f | D) &= \frac{1}{P(D)} \sum_\prec P(\! \prec) P(f, D | \! \prec)\\
\end{split}
\label{eq:modularSTL}
\end{equation}
where $\pi_i = \pi(X_i)$ is the parent set of variable $i$. The unconditional posterior for the features is obtained by summing over orders, using the following steps:
\begin{enumerate}
\item Calculate family scores: $\beta_i(\pi_i) = P(\pi_i) P(x_i | \pi_i) f_i(\pi_i)$ for each node $i$ and potential set of parents $\pi_i$. The computational complexity of each of these is some function $C(m)$ of the number of samples $m$. The maximum number of parents allowed for any node is typically fixed to a small natural number, $r$. Therefore, there are $O(N^{r+1})$ of these functions to calculate for a total complexity of $O(N^{r+1} C(m))$.

\item Calculate local contribution of each subset $U \subseteq V - \{i\}$ of potential parents of $i$: $\alpha_i(U) = \sum_{\pi_i \subseteq U} P(\pi_i) P(x_i | \pi_i) f_i(\pi_i)$. Using a truncated fast M\"{o}bius transform and pre-computed $\beta$'s, all of the $\alpha$ functions are computed in $O(n2^n)$ time.

\item Sum over the subset lattice of the various $U_i$ to obtain the sum over orders $\prec$. Using dynamic programming, this sum takes time $O(n2^n)$.
\end{enumerate}
The total computational complexity for a single task is $O(n2^n + n^{r+1}C(m))$. This is the exact calculation of the posterior. For large networks, roughly $n > 30$, the exponential term is intractible. In these cases, MCMC simulations give an approximation to the posterior probability, so that $P(f|D) \approx \frac{1}{T} \sum_{t=1}^T P(\! \prec_t) P(f | D, \! \prec_t)$ for $\! \prec_t$ sampled from order space \cite{friedman_being_2003} or partial orders \cite{Niinimaki11}.
\label{sec:koivistoAlg}


\subsection{Multitask Bayesian Networks}

Multitask Bayesian network learning leverages knowledge among a set of related tasks by applying a bias toward learning similar networks among the tasks. The underlying assumption is that much of the network is shared among tasks, yet a few differences may exist. By leveraging information among tasks, we can learn more robust networks than would be possible from a single small sample \cite{niculescu-mizil_inductive_2007}. We will apply a similar bias mechanism for network discovery. First we describe the objective function of existing MAP estimate algorithms, which has been shown to be effective at leveraging information. A set of tasks with data sets $D^{(k)}$ and networks $G^{(k)}$ for $k \in \{1,\ldots, K\}$ can be learned by optimizing:
\begin{equation}
\begin{split}
&P(\mathcal{G} | \mathcal{D}) = P(G^{(1)}, \ldots, G^{(K)} | D^{(1)}, \ldots, D^{(K)} ) =  \\
	&P(D^{(1)}, \ldots, D^{(K)} | G^{(1)}, \ldots, G^{(K)}) \frac{P(\mathcal{G})} {P(\mathcal{D})}
\end{split}
\label{eq:posteriorGeneral}
\end{equation}
In existing multitask network learning formulations, the joint structure prior, $P(\mathcal{G})$, is used to encode a bias toward similar structures by penalizing differences in network structure among tasks \cite{niculescu-mizil_inductive_2007,oyen_aaai_2012}. We can assume that $P(D^{(k)} | G^{(k)})$ is independent of all other $G^{(i)}$ so Eq~\ref{eq:posteriorGeneral} simplifies and the joint prior over structures can be described by pairwise sharing of information among tasks as in Eq~\ref{eq:posteriorMTL}.
\begin{equation}
\begin{split}
&P_{\mathit{MTL}}(\mathcal{G} | \mathcal{D}, \lambda) = \frac{P(\mathcal{G} | \lambda)}{P(\mathcal{D})} \prod_{k=1}^K P(D^{(k)} | G^{(k)})\\
&P(\mathcal{G} | \lambda) = \frac{1}{Z} \prod_{k=1}^K P(G^{(k)}) \prod_{i=1}^{k-1} \lambda (1 - \lambda)^{\Delta(G^{(k)}, G^{(i)})}
\end{split}
\label{eq:posteriorMTL}
\end{equation}
where $Z$ is a normalization constant and $\Delta$ is any graph distance metric, such as edit distance, that measures the number of structural differences between graphs $G^{(k)}$ and $G^{(i)}$. 

\section{Multitask Feature Discovery}

In this section, we present our novel Bayesian approach to structure discovery in multitask Bayesian networks. The challenge is finding a way to bias structures to be similar among tasks, like Eq~\ref{eq:posteriorMTL}, while maintaining the efficiency of calculating feature posteriors that factor into local calculations, like Eq~\ref{eq:modularSTL}. First we formulate the problem, describe structural bias terms that are order-modular, provide a Bayesian approach for handling the strength of the bias, and finally describe practical implementation issues.

\subsection{Problem Formulation}

Instead of learning the feature posteriors from a single task-specific data set, we have $K$ tasks from which we will leverage data. We define the indicator $f^{(k)} = f(G^{(k)})$. Our goal is to learn a feature for each task $P(f^{(k)} | D^{(1)}, \ldots, D^{(K)})$ $\forall k \in \{1, \ldots, K\}$. Again, all formulations are written for a single feature (e.g. a directed edge), but calculating them simultaneously (e.g. all edges in a network) takes the same time. To simplify the development of the objective, we will consider, without loss of generality, the case where $K=2$.
\begin{equation}
\begin{split}
P&(f^{(1)} | D^{(1)}, D^{(2)}) = \\
& = \sum_{\prec} P(\prec) \!\!\!\! \sum_{G^{(1)} \subseteq \prec}\!\!\!\! P(G^{(1)} | D^{(1)}, D^{(2)}) f(G^{(1)}) \\
&= \sum_{\prec}  \sum_{G^{(1)} \subseteq \prec} \!\!\!\! f(G^{(1)}) \times \! \left[ \sum_{G^{(2)} \subseteq \prec}\! P(G^{(1)}, G^{(2)} | D^{(1)}, D^{(2)}) \right] \\
&= \frac{1}{P(D^{(1)}, D^{(2)})} \sum_\prec \sum_{G^{(1)} \subseteq \prec}\! f(G^{(1)}) P(D^{(1)} | G^{(1)}) \times \\
	&\left[ \sum_{G^{(2)} \subseteq \prec}\!\!\!\! P(D^{(2)} | G^{(2)}) P(G^{(1)}, G^{(2)}) \right] \\
\end{split}
\label{eq:formulation}
\end{equation}

For the purpose of calculating the transfer bias, we impose the same order, $\prec$, on both tasks, and then marginalize over orders. This restriction makes computation more efficient, and it seems reasonable to bias a feature contingent on a particular ordering toward the evidence from other tasks while they are restricted to the same set of possible graph structures. Rewriting Eq~\ref{eq:formulation} as a joint probability conditioned on an order, we get the following:
\begin{equation}
\begin{split}
P(f^{(1)}&, D^{(1)}, D^{(2)} | \! \prec) = \sum_{G^{(1)} \subseteq \! \prec} \!\!\!\! f(G^{(1)}) P(D^{(1)} | G^{(1)}) \times \\
&\left[\sum_{G^{(2)} \subseteq \! \prec} \!\!\!\! P(D^{(2)} | G^{(2)}) P(G^{(1)}, G^{(2)} | \! \prec) \right]
\end{split}
\label{eq:conditionalPosterior}
\end{equation}
Our formulation imposes a transfer bias at the level of structure, $P(G^{(1)}, G^{(2)} | \! \prec)$, which is more intuitive than at the level of orders. To calculate this sum efficiently, it is necessary to factor it into a product over local sums. We prove that this is indeed possible, for appropriately chosen structure priors. In addition to the modularity assumptions already stated, we impose an additional modularity assumption, which we call Assumption 4) \textbf{Transfer prior modularity}: $P(G^{(1)}, G^{(2)} | \! \prec) = \prod_{i=1}^n P(\pi_i^{(1)}, \pi_i^{(2)} | U_i)$. Examples of priors that obey this assumption are graph distance measures that count the number of edge additions and deletions, so this is a reasonable requirement.

\begin{theorem}
If $G^{(1)}, \ldots, G^{(K)}$ obey the four assumptions of modularity, then
\begin{equation*}
\begin{split}
P(f^{(1)}, \mathcal{D} &| \! \prec)
	= \prod_{i \in V} \sum_{\pi_i^{(1)} \subseteq U_i} 
	f_i(\pi_i^{(1)}) P(x_i^{(1)} | \pi_i^{(1)}) \times \\
& \left[\sum_{\pi_i^{(2)} \subseteq U_i} P(x_i^{(2)} | \pi_i^{(2)}) P(\pi_i^{(1)}, \pi_i^{(2)} | U_i) \right]\\
\end{split}
\end{equation*}
\label{thm:modularObjective}
\end{theorem}
\begin{proofsketch}
Apply the chain rule and marginalize over graph structure to get Eq~\ref{eq:conditionalPosterior}. Use the modularity properties on each term in the product, and notice that the result factors into the desired form.
\begin{equation*}
\begin{split}
P&(f^{(1)}, \mathcal{D} | \! \prec) = 
	\sum_{G^{(1)} \subseteq \! \prec} f(G^{(1)}) P(D^{(1)} | G^{(1)}) \times \\
&\left[ \sum_{G^{(2)} \subseteq \! \prec} P(D^{(2)} | G^{(2)}) P(G^{(1)}, G^{(2)} | \! \prec) \right]\\
= &\sum_{\pi_1^{(1)} \subseteq U_1} \cdots \sum_{\pi_n^{(1)} \subseteq U_n} \prod_{i=1}^n
	\left[ f_i(\pi_i^{(1)}) P(x_i^{(1)} | \pi_i^{(1)}) \right] \times \\
& \left[ \sum_{\pi_1^{(2)} \subseteq U_1} \cdots \sum_{\pi_n^{(2)} \subseteq U_n} \prod_{i=1}^n
	P(x_i^{(2)} | \pi_i^{(2)}) P(\pi_i^{(1)}, \pi_i^{(2)} | U_i) \right] \\
\end{split}
\end{equation*}
\end{proofsketch}

The details of the proof are straightforward, yet space consuming, and so are omitted. See \cite{Koivisto:2004:EBS:1005332.1005352} for a similar proof.

\subsection{Computational Complexity}
\label{sec:complexity}

The power of Theorem~\ref{thm:modularObjective} is the computational savings that we gain. Using the factored posterior, we only need to change Step 1 of the order-space algorithm outlined in Section~\ref{sec:koivistoAlg}, the calculation of the family scores. The transfer-biased family scores are calculated as:
\begin{equation}
\begin{split}
\beta_{ki}(\pi_i) &= f_i(\pi_i^{(k)}) P(x_i^{(k)} | \pi_i^{(k)}) P(\pi_i^{(k)}, \pi_i^{(j)})\\
	&= f_i(\pi_i^{(k)}) P(x_i^{(k)} | \pi_i^{(k)}) \times \\
	&\left[\sum_{j \neq k} \sum_{\pi_i^{(j)} \subseteq U_i} P(x_i^{(j)} | \pi_i^{(j)}) 
	P(\pi_i^{(k)}, \pi_i^{(j)}) \right]
\end{split}
\end{equation}
There are now $Kn^{r+1}$ of these families to calculate and each one has a sum over $O(Kn^r)$ terms. The computational complexity increases from the single-task time of $O(n^{r+1} C(m))$ to $O(K^2 n^{2r+1} C(m))$. Steps 2 and 3 remain unchanged with an exponential complexity that can be reduced through MCMC approximation.

Even the polynomial term becomes unmanageable for networks with more than 30 or so nodes and must be approximated. We note that in many cases the family scores $P(x_i^{(j)} | \pi_i^{(j)})$ are exponentially larger for some $\pi_i^{(j)} \subseteq U_i$ than others. Therefore, we can use a simple approximation by summing over only the most likely parent sets. While calculating the family scores, we create a set of the highest-scoring families, called set $\mathcal{H}_i^{(k)}$ for each node in each task. To populate this set, we simply include $h$ parent sets that give the highest $P(x_i^{(k)} | \pi_i^{(k)})$, for some constant $h$. Then we use the approximate structural prior:
\begin{equation*}
P_i(\pi_i^{(k)}) \approx \sum_{\pi_i^{(j)} \in \mathcal{H}_i^{(j)}} P(x_i^{(j)} | \pi_i^{(j)}) P(\pi_i^{(k)}, \pi_i^{(j)} | U_i)
\end{equation*}

\subsection{Transfer via Structure Bias}

Now that we know we can incorporate transfer bias, we need to select a modular bias term that transfers knowledge among tasks. The MAP multitask algorithms use a penalty on the number of differences between tasks using a graph distance function. In that case, the number of edges that must be added, deleted, or reversed to edit one graph into the other is penalized. Due to our modularity constraint, our transfer bias must be defined as a function on pairs of parent sets $(\pi_i^{(k)}, \pi_i^{(j)})$, rather than graphs. We choose to penalize the number of edge additions which breaks down into local calculations: the number of parents present in $\pi_i^{(k)}$ that are not present in $\pi_i^{(j)}$. In other words, the size of the set difference $\Delta_{ikj} = \abs{\pi_i^{(k)} \setminus \pi_i^{(j)}}$ will be biased toward small values. To encourage the number of differences to be small, we apply a penalty in the form of a geometric distribution,
\begin{equation}
P(\pi_i^{(k)}, \pi_i^{(j)} | U_i, \lambda) = \frac{1}{Z} (1-\lambda) ^ {\Delta_{ikj}}
\label{eq:structurePrior}
\end{equation}
Calculation of the normalization constant requires summing over an exponential number of possible combinations of parent sets $(\pi_i^{(k)}, \pi_i^{(j)})$. However, we found show how to simplify the sum into an easy closed form. We employ a shortcut by noting that there are a finite number of values that $\Delta_{ikj}$ can take and we find a closed form for calculating the number of parent-set combinations that produce each value of $\Delta_{ikj}$. 
\begin{equation}
\begin{split}
Z &= \sum_{\pi_i^{(k)} \subseteq U_i} \sum_{\pi_i^{(j)} \subseteq U_i} (1-\lambda)^{\Delta_{ikj}} \\
	&= (4 - \lambda)^{\abs{ U_i }} \\
\end{split}
\label{eq:structurePriorNorm}
\end{equation}
Here we give a sketch of the derivation of Eq~\ref{eq:structurePriorNorm}. First, we simplify the inner sum by fixing parent set $\pi_i^{(1)}$ and counting how many parent sets $\pi_i^{(2)}$ will give $\Delta_{ikj} = 0$ ($\pi_i^{(2)}$ can contain any parents from the set $\{U_i \setminus \pi_i^{(1)}\}$ but \emph{none} from $\pi_i^{(1)}$); then how many $\pi_i^{(2)}$ will give $\Delta_{ikj} = 1$ ($\pi_i^{(2)}$ can contain any parents from the set $\{U_i \setminus \pi_i^{(1)}\}$ and \emph{exactly one} from $\pi_i^{(1)}$); etc, up to the maximum of $\Delta_{ikj} = \abs{\pi_i^{(1)}}$. This sum turns out to be a binomial expansion, and so we can write it in closed form. Next, we perform a similar expansion of the outer sum over parent sets $\pi_i^{(1)}$ that have size $\abs{\pi_i^{(1)}} = 0$, and $\abs{\pi_i^{(1)}} = 1$, etc up to the maximum $\abs{\pi_i^{(1)}} = \abs{U_i}$. This sum also turns out to be a binomial expansion and therefore can be simplified into a closed form.\footnote{Algebraic details of this calculation are space consuming and can be supplied in an online supplement.}

Plugging Eq~\ref{eq:structurePriorNorm} into Eq~\ref{eq:structurePrior} gives the structure prior:
\begin{equation}
\begin{split}
P(\pi_i^{(k)}, \pi_i^{(j)} | U_i, \lambda) &=  \frac{(1-\lambda) ^ {\Delta_{ikj}}} {(4 - \lambda)^{\abs{U_i}}} \\
\end{split}
\end{equation}
The parameter $\lambda$, $0 \leq \lambda \leq 1$, controls the strength of transfer bias. When $\lambda = 0$, the prior becomes uniform and therefore there is no transfer. When $\lambda = 1$, the prior is non-zero only when no edge additions occur, and therefore the only parents that are allowed are those that are likely in the other tasks.

\subsection{Bayesian Model Averaging}

We have just introduced an additional parameter, $\lambda$, which is a bit of a nuisance. Existing MAP algorithms cannot avoid dealing with this, and they typically estimate $\lambda$ by optimizing over a held out validation set. This is computationally expensive and reduces the amount of available data for training. Rather than selecting a fixed value for $\lambda$, we perform Bayesian model averaging over all possible values of $\lambda$. This Bayesian approach is compelling as the true amount of similarity among tasks is unknown, and the ``true" value of $\lambda$ is only incidental to our objective of learning the structure likelihoods. Furthermore, it saves us the computation of running the algorithm for several values of $\lambda$, and we do not need to hold-out data for tuning.

We set an uninformative uniform prior, $p(\lambda | U_i) = 1$ for $0 \leq \lambda \leq 1$, and  marginalize over $\lambda$.
\begin{equation}
\begin{split}
P(\pi_i^{(k)}, \pi_i^{(j)} | U_i) &=  \int_0^1 P(\pi_i^{(k)}, \pi_i^{(j)} | U_i, \lambda) p(\lambda | U_i) d\lambda \\
&=  \int_0^1 \frac{(1-\lambda) ^ { \Delta_{ikj}}} {(4 - \lambda)^{\abs{U_i}}} d\lambda \\
&= \frac{_2\text{F}_1(\abs{U_i}, 1; \Delta_{ikj}+2; 1/4)}{4^{\abs{U_i}} (\Delta_{ikj} + 1)}\\
\end{split}
\label{eq:biasTerm}
\end{equation}
where $_2\text{F}_1$ is the ordinary hypergeometric function:
\begin{equation*}
\begin{split}
_2\text{F}&_1(\abs{U_i}, 1; \Delta_{ikj}+2; 1/4) = \\
 & \sum_{n=0}^\infty \frac{\Gamma(1+n)}{\Gamma(1)} \cdot
	\frac{\Gamma(\abs{U_i} + n)}{\Gamma(\abs{U_i})}  \cdot
	\frac{\Gamma(\Delta_{ikj} + 2)}{\Gamma(\Delta_{ikj} + 2 + n)}  \cdot \frac{1}{4^nn!} \\
\end{split}
\end{equation*}

The last step in Eq~\ref{eq:biasTerm} is obtained by applying an identity given by Euler in 1748 \cite{Bailey1935Generalised-Hyp}. If $\beta$ is the beta function and $_2\text{F}_1$ is the ordinary hypergeometric function, then
\begin{equation*}
\begin{split}
\int_0^1 x^{b-1}(1-x)^{c-b-1}&(1-zx)^{-a} dx = \\
&\beta(b, c-b) _2\text{F}_1 (a, b; c; z)
\end{split}
\end{equation*}
for $\Re(c) > \Re(b) > 0$. We let $x = \lambda$, $a = \abs{U_i}$, $b = 1$, $c = \Delta_{ikj} + 2$, and $z = 1/4$. Then the condition, $\Delta_{ikj} + 2 > 1 > 0$, holds for any $\Delta_{ikj} \geq 0$ which is the valid range for $\Delta_{ikj}$. Plugging these values into the identity gives the solution to the integral as:
\begin{equation*}
\beta(1, \Delta_{ikj} + 1) _2\text{F}_1 (\abs{U_i}, 1; \Delta_{ikj} + 2; 1/4)
\end{equation*}
which simplies to the solution given in Eq~\ref{eq:biasTerm}.

We are only interested in calculating $_2\text{F}_1$ for combinations of integer-values of $\Delta_{ikj}$ and $\abs{U_i}$ for $0 \leq \Delta_{ikj} \leq \abs{U_i} < n$. For these values, $_2\text{F}_1$ is convergent and efficient solvers exist. Thus, we can plug the result of Eq~\ref{eq:biasTerm} into the equation of Theorem~\ref{thm:modularObjective}.

Bayesian model averaging is made possible by conditioning on orders. Existing multitask network learning algorithms that search in DAG space \cite{niculescu-mizil_inductive_2007,oyen_aaai_2012} would be required to calculate a normalization constant like that in Eq~\ref{eq:structurePriorNorm} but with sums over all possible DAGs, and no closed form has been found for such a sum.

\begin{figure}[tb]
\centering
\includegraphics[width=0.6\columnwidth]{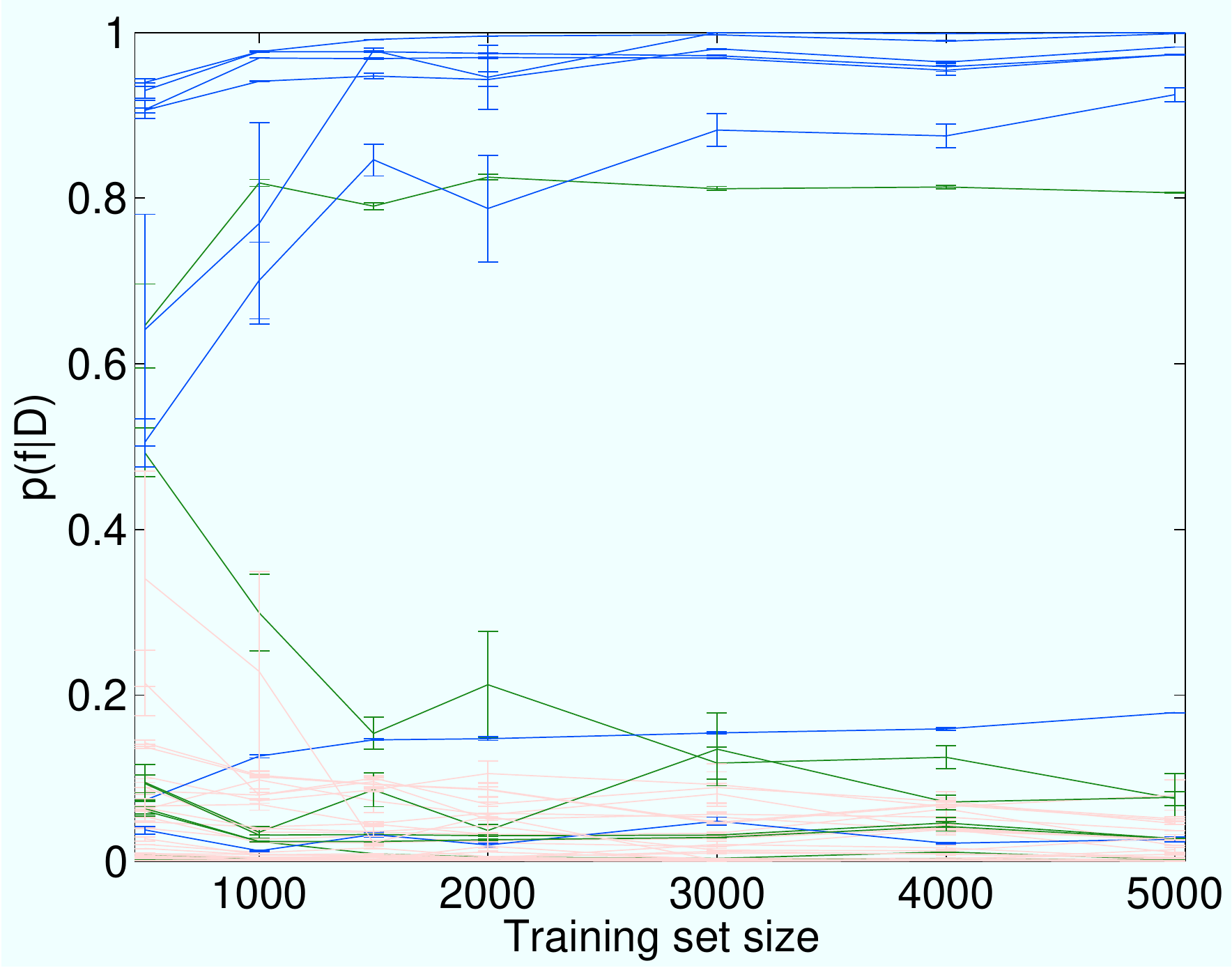}
\caption{Posterior probability estimate for each edge in the \textit{asia} network from various large sample sets (means calculated from 20 sample sets). Blue curves are true edges, green are reverse of a true edge, pink are non-edges. (Best viewed in color.)}
\label{fig:asiaLargeSamp}
\end{figure}

\section{Experiments}

Multitask learning should be able to identify true edges and non-edges with less data than is possible with traditional single-task learning. We compare our MTL structure discovery algorithm against two baselines. The first baseline is single-task learning (STL), where each network is learned independently of the other tasks. The other baseline (POOL) takes the opposite extreme by pooling data from all tasks together and treating it as a single task. POOL uses the strongest leveraging of data among tasks possible and so it should perform best if the separate tasks are actually the same distribution.  For all approaches, we use the BeanDisco implementation for exact and approximate network discovery \cite{Niinimaki11}. For MTL, the scoring function of BeanDisco is modified as described above. For POOL, the data are merged before applying the algorithm. 

\subsection{Benchmark Data}

\begin{figure*}[htb]
\centering
\subfloat[STL]
	{\includegraphics[width=0.32\textwidth]{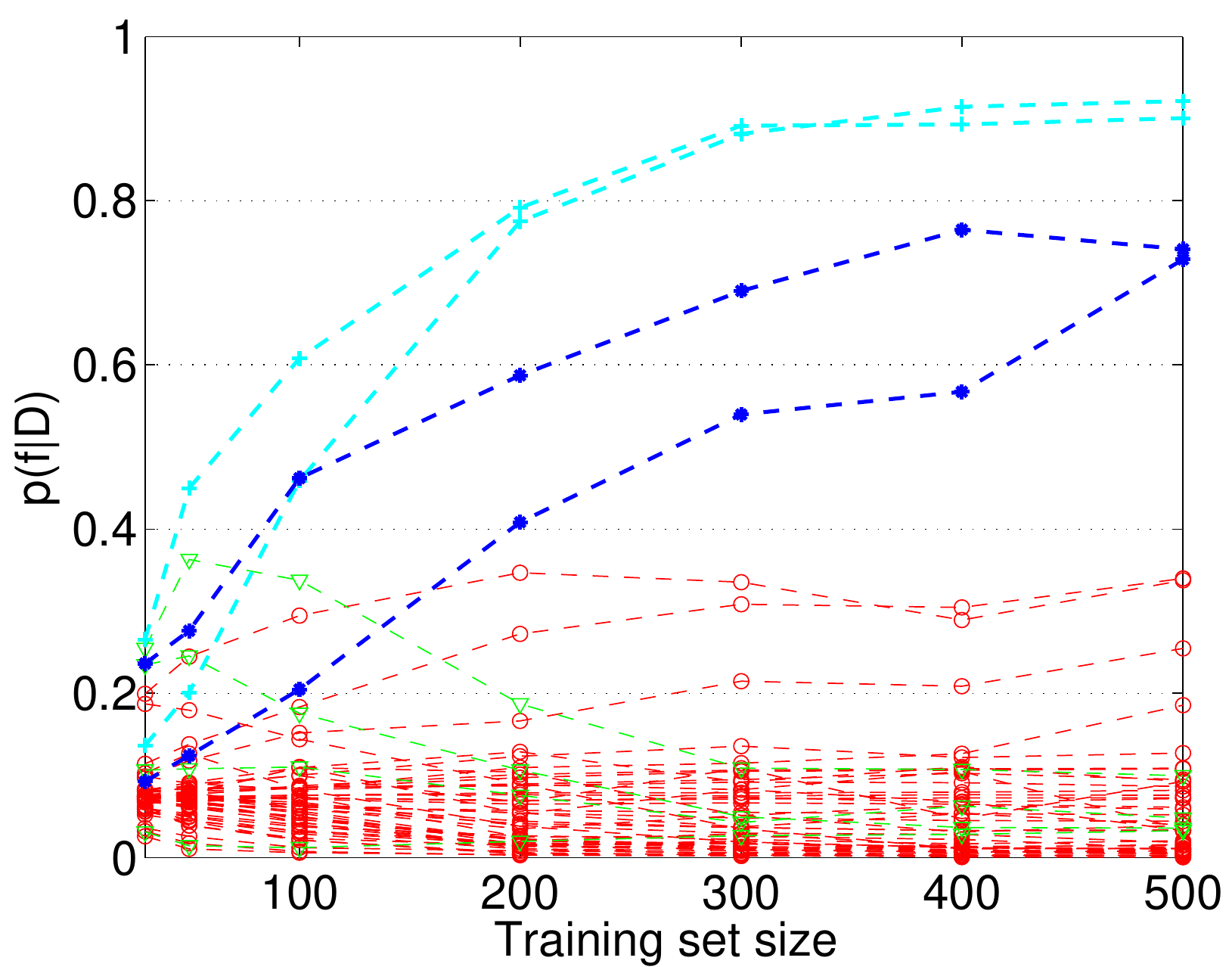}
	\label{fig:asiaPedgeSTL}}
	\hfil
\subfloat[MTL]
	{\includegraphics[width=0.32\textwidth]{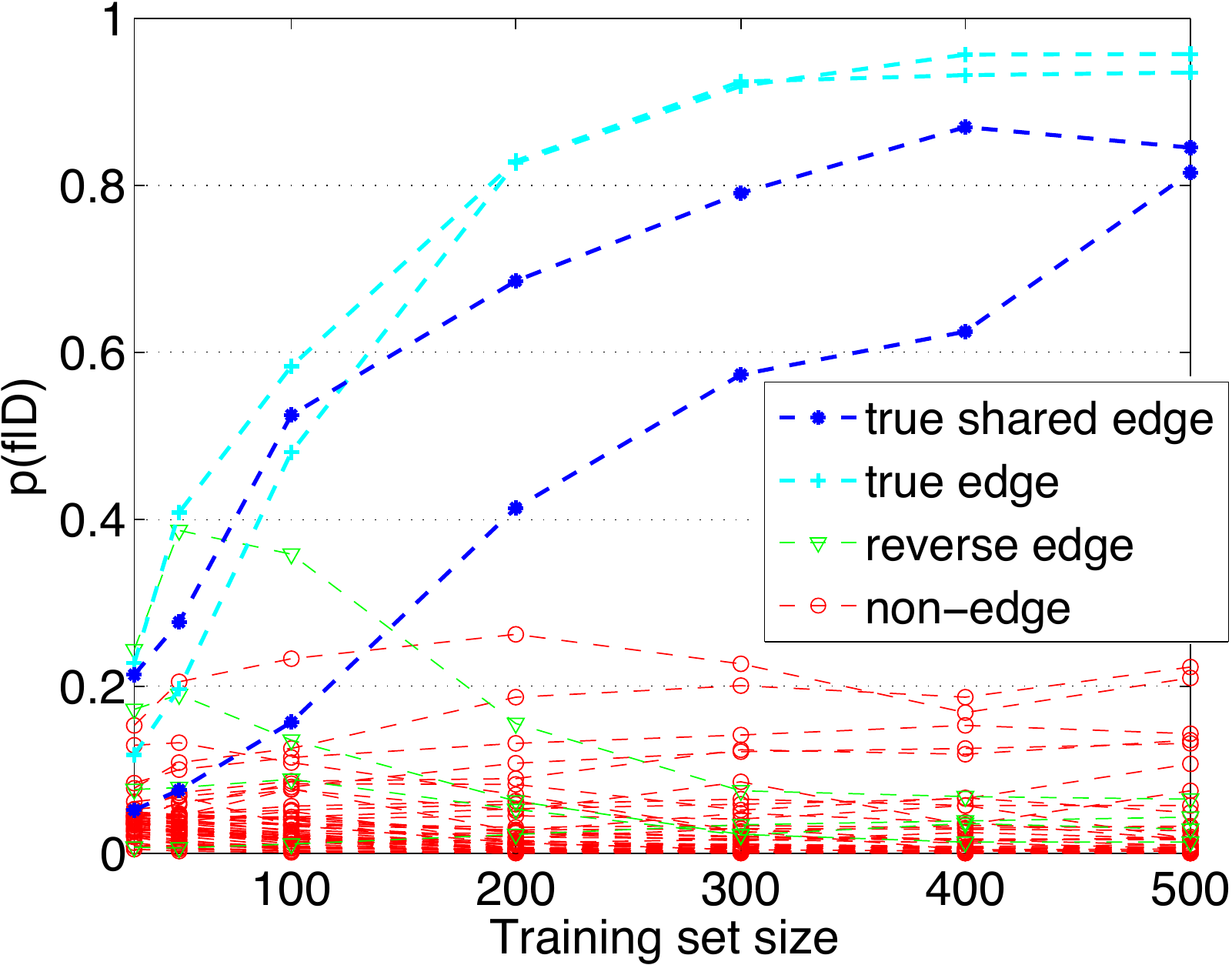}
	\label{fig:asiaPedgeMTL}}
	\hfil
\subfloat[POOL]
	{\includegraphics[width=0.32\textwidth]{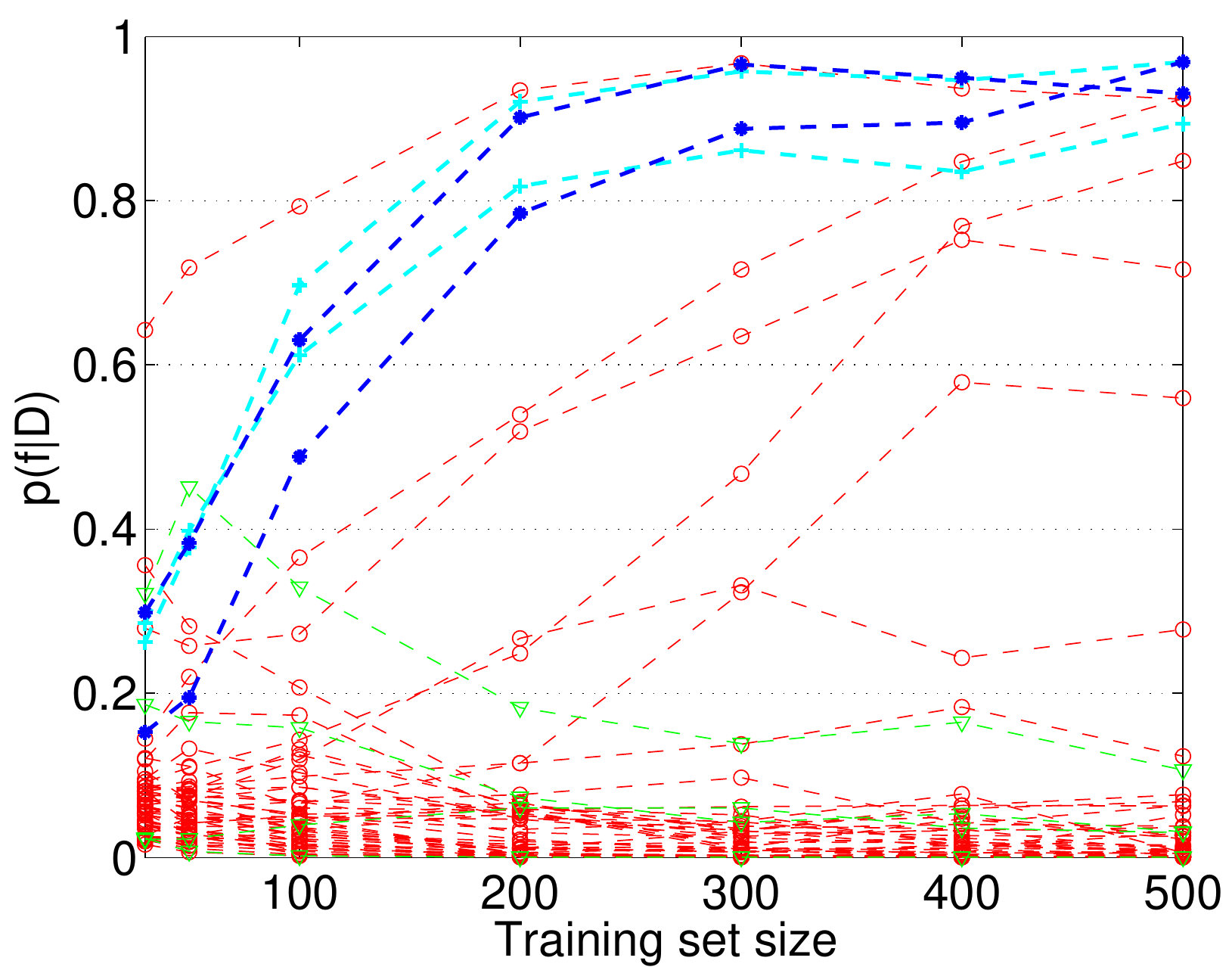}
	\label{fig:asiaPedgePOOL}}
\caption{Example posterior probability estimate for each edge in a modified \textit{asia} network from various small sample sets (means calculated from 20 sample sets). Up is good for blue and cyan curves. Down is good for red and green curves.}
\label{fig:asiaPedge}
\end{figure*}

Synthetic data is generated from benchmark Bayesian networks, \textit{asia} which has 8 variables \cite{lauritzen1988local} and \textit{alarm} which has 37 variables \cite{beinlich89alarm}. Even when the generative model is known, it is not obvious how to measure the performance of a network discovery algorithm. To give a clear picture of our objective, see Figure~\ref{fig:asiaLargeSamp}. On the small \textit{asia} network we can calculate the true posterior likelihoods of structural features given the data. Sample noise affects the true posterior likelihood of structural features but the posteriors appear to stabilize for large training sets. Even so, one edge has been consistently identified in the reverse direction of the true edge and another true edge represents such a subtle dependency that it is not discernible from this amount of data. Therefore, we use the posterior estimates from large training sets as our ground truth $P^*(f|D)$. In the case of \emph{asia}, $P^*(f|D) = \hat{P}_{\mathit{STL}}(f|D_{5000})$.

We need a set of related networks and so we modify some structures of the given benchmark network to create similar but different networks. We delete each edge with some probability $p_{\mathit{del}}$ and vary $p_{\mathit{del}}$ from 0.1 to 0.5 to create sets of networks with more or less features in common for various experiments. If an edge is deleted, the conditional probability table for the child of the deleted edge is updated by marginalizing over the deleted parent. In our experiments, the full generative model is repeated 10 times to produce 10 different sets of $K$ networks each for a given $p_{\mathit{del}}$.

\subsection{Benchmark Results}

The goal of transfer learning is to accelerate the learning curve at smaller training set sizes by leveraging data among similar tasks. If we look closely at the results for one particular modified \textit{asia} network, we can see what effect multitask learning has. Figure~\ref{fig:asiaPedgeSTL} shows the estimated posteriors from STL at smaller sample sizes. Even at these small sample sizes, the posteriors of the true edges tend to be higher than those of non-edges. However, compared to the estimates from large samples, these posteriors exhibit high variance and many are quite far from the large-sample posterior value (in the figure, error bars omitted for readability). The question is whether multitask learning can produce a steeper learning curve.

Figure~\ref{fig:asiaPedgeMTL} shows the learning curve achieved by MTL on the same network, using data leveraged from one other task, where some but not all edges are in common between the two tasks. This learning curve shows a wider gap between the estimated values of true edges and non-edges. In that sense, the learning curve is better than STL because it is better at separating the true edges from the non-edges at smaller training set sizes. In particular, this gain is achieved through the lower estimates of non-edges. In other words, non-edges are more quickly identified as such through transfer learning than without. On the other hand, the raw estimates of true edges tend to have such high variance (both with STL and with MTL) that it is not possible to say that one algorithm is doing better than the other in terms of converging on the actual $P^*(f|D)$. 

Figure~\ref{fig:asiaPedgePOOL} shows the learning curve obtained by POOL on the same modified \textit{asia} network. POOL combined the data from two modified networks with some edges in common. The algorithm effectively has twice as much data to work with as STL, therefore the learning curves are steeper. Yet there are quite a few non-edges with high posterior values.

\begin{table}[tb]
\centering
\caption{Performance increase for \emph{asia} in terms of AUC given by MTL vs STL and MTL vs POOL}
\label{table:auc}
\begin{footnotesize}
\begin{tabular}{c || r | c || r | c}
  Training & \multicolumn{2}{c||}{MTL vs STL} & \multicolumn{2}{c}{MTL vs POOL} \\
  samples & \% incr & pair-t & \% incr & pair-t \\
  \hline
  5 & 3.06 & - & 10.72 & MTL \\
  10 & 9.02 & MTL & 4.10 & - \\
  20 & 4.90 & MTL & 0.34 & - \\
  30 & 4.98 & MTL & 0.85 & - \\
  40 & 7.60 & MTL & 3.66 & - \\
  50 & 3.00 & MTL & 3.08 & MTL \\
  100 &  1.97 & MTL & 2.96 & MTL \\
  200 & 0.53 & - & 2.82 & MTL \\
  400 & 0.14 & - & 4.14 & MTL \\
  500 & -0.03 & - & 3.72 & MTL \\
 \end{tabular}
 \end{footnotesize}
\end{table}

\begin{figure}[tb]
\centering
\subfloat[10 samples per task]
	{\includegraphics[width=0.48\columnwidth]{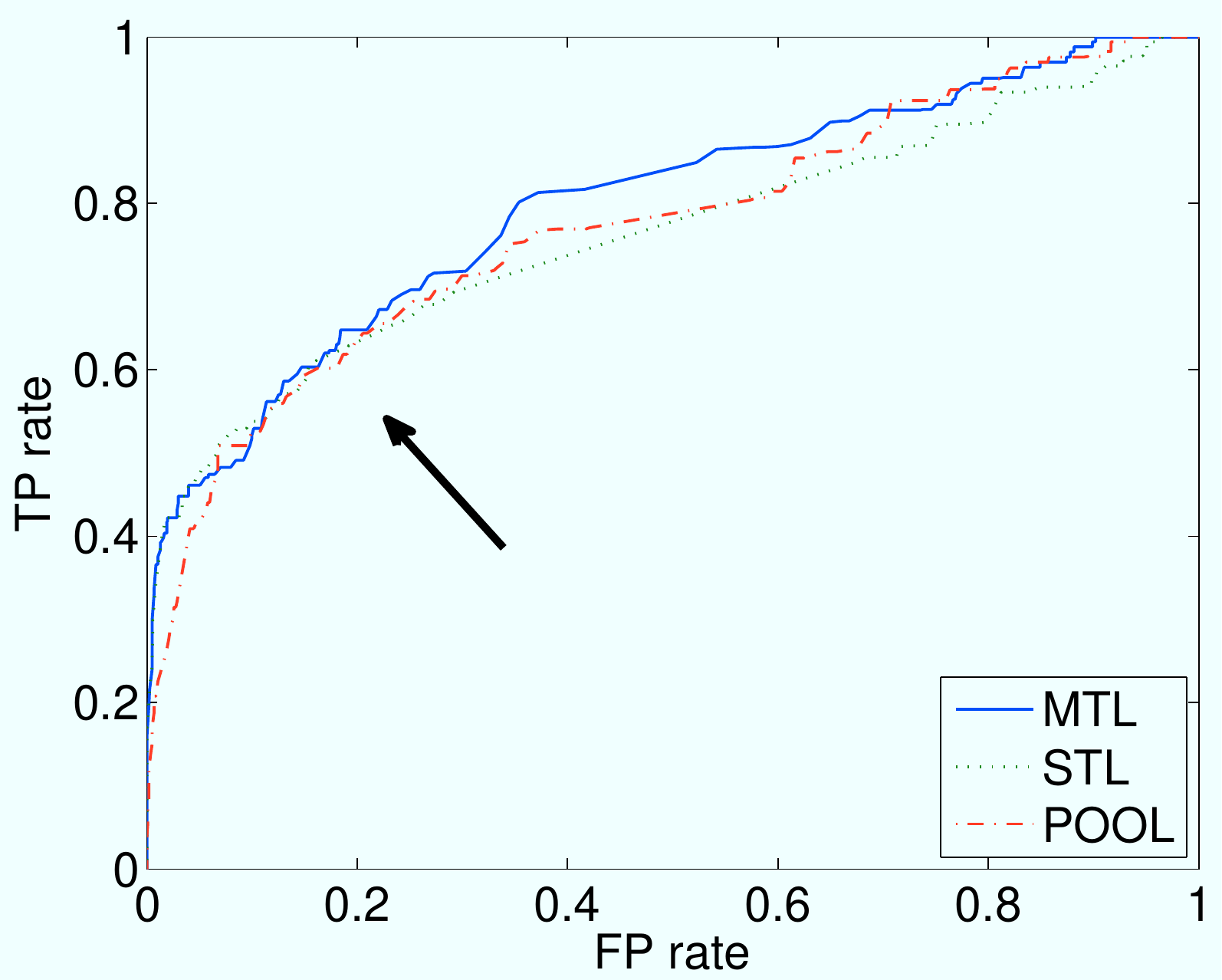}
	\label{fig:asiaRoc10}}
	\hfil
\subfloat[50 samples per task]
	{\includegraphics[width=0.48\columnwidth]{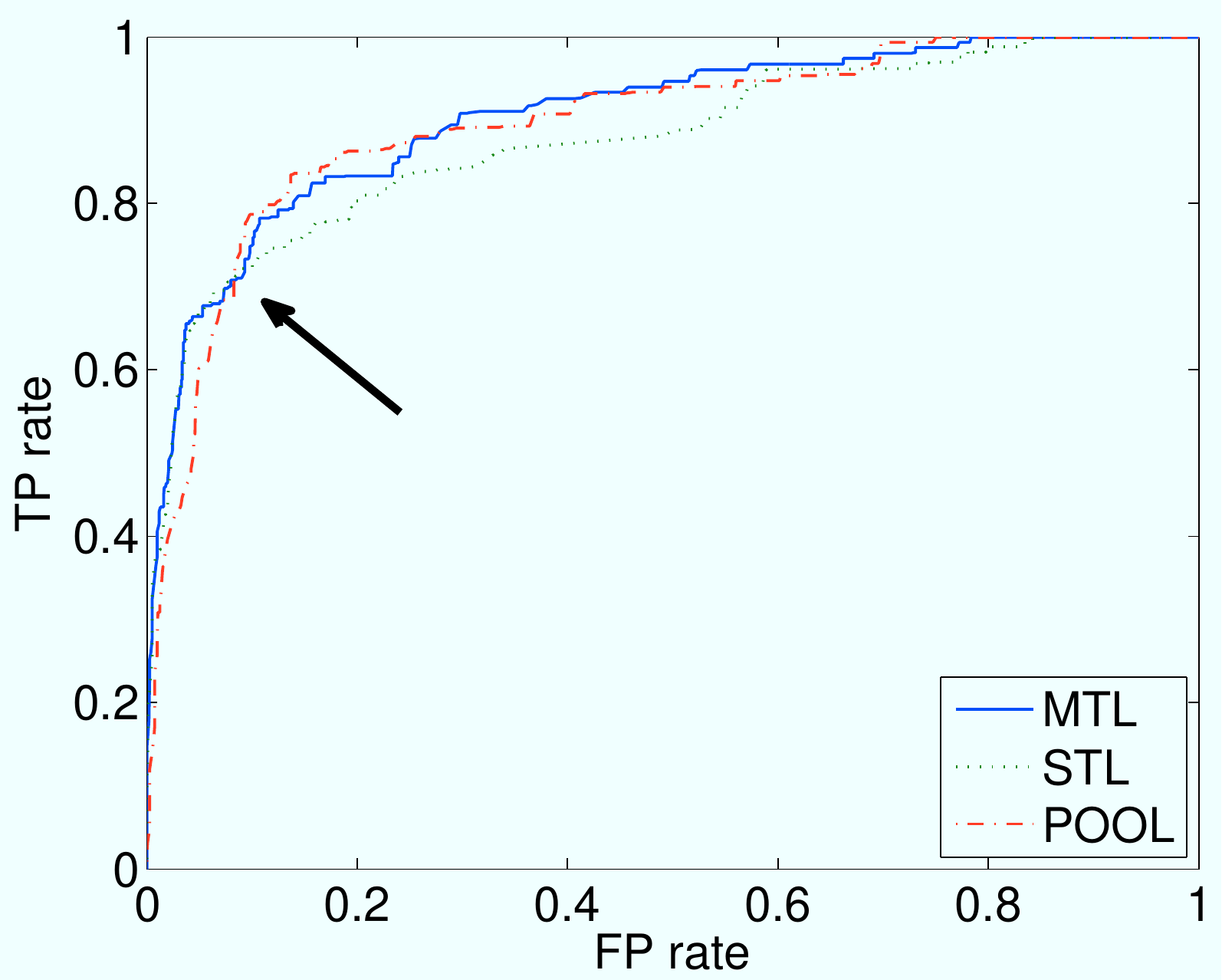}
	\label{fig:asiaRoc50}}
\caption{ROC curves for \textit{asia}. Each point is the (FP rate, TP rate) aggregated over 2 tasks and 10 trials of the generative model, for a particular value of $\tau$. Arrow indicates where STL and POOL curves cross.}
\label{fig:asiaRoc}
\end{figure}

To quantify these results, we measure how well the estimates for true edges separate from the estimates for non-edges. There is potentially a directed edge between each ordered pair of nodes. We call this set of potential edges $E$. We quantify the ground truth using the $P^*(f|D)$ values obtained from large samples and identify the set of ``true" edges $E^* = \{f \in E \: | \: P^*(f|D) > 0.5\}$. 


To differentiate learned edges from learned non-edges, we assign a threshold $\tau$ and call any feature an edge if its posterior is greater than the threshold, $\hat{E} = \{f \in E | \hat{P}(f|D) > \tau\}$. By varying $\tau$, $0\leq \tau \leq 1$, we can investigate the tradeoff between the rates of true-positives (TP) and false-positives (FP) in $\hat{E}$ by constructing an ROC curve (Figure~\ref{fig:asiaRoc}). The TP rate is $\abs{\hat{E} \cap E^*} / \abs{\hat{E}}$. The FP rate is $\abs{\hat{E} \setminus E^*} / (\abs{E} - \abs{\hat{E}})$.

Figure~\ref{fig:asiaRoc} shows that various algorithms have different strengths along the ROC curve. The ROC shows that the patterns indicated in Figure~\ref{fig:asiaPedge} are borne out more generally; that is, STL is slowest to positively identify true edges, while POOL has difficulty eliminating false positives. MTL achieves the greatest overall separation of true edges and non-edges. Initially, at the left end of the ROC curve, with low false positive rates both MTL and STL perform best, finding more true positives than POOL. However, the performance of STL falls off as the false positive rate increases: STL is missing some true positives that both MTL and POOL are able to identify. MTL gives us the best of both worlds, giving the best overall performance.

\begin{figure*}[tb]
\centering
\subfloat[STL]
	{\includegraphics[width=0.32\textwidth]{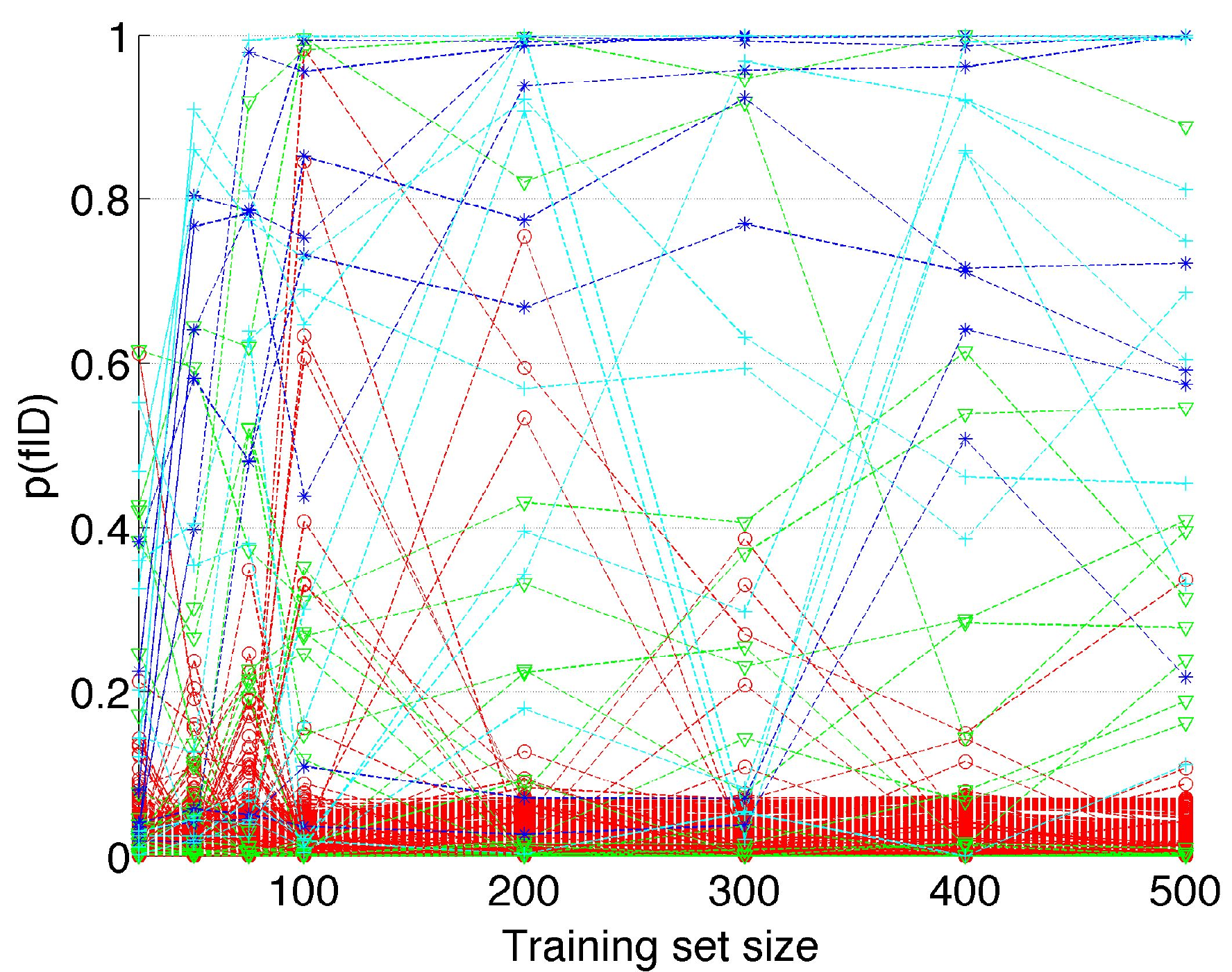}
	\label{fig:alarmPedgeSTL}}
	\hfil
\subfloat[MTL]
	{\includegraphics[width=0.32\textwidth]{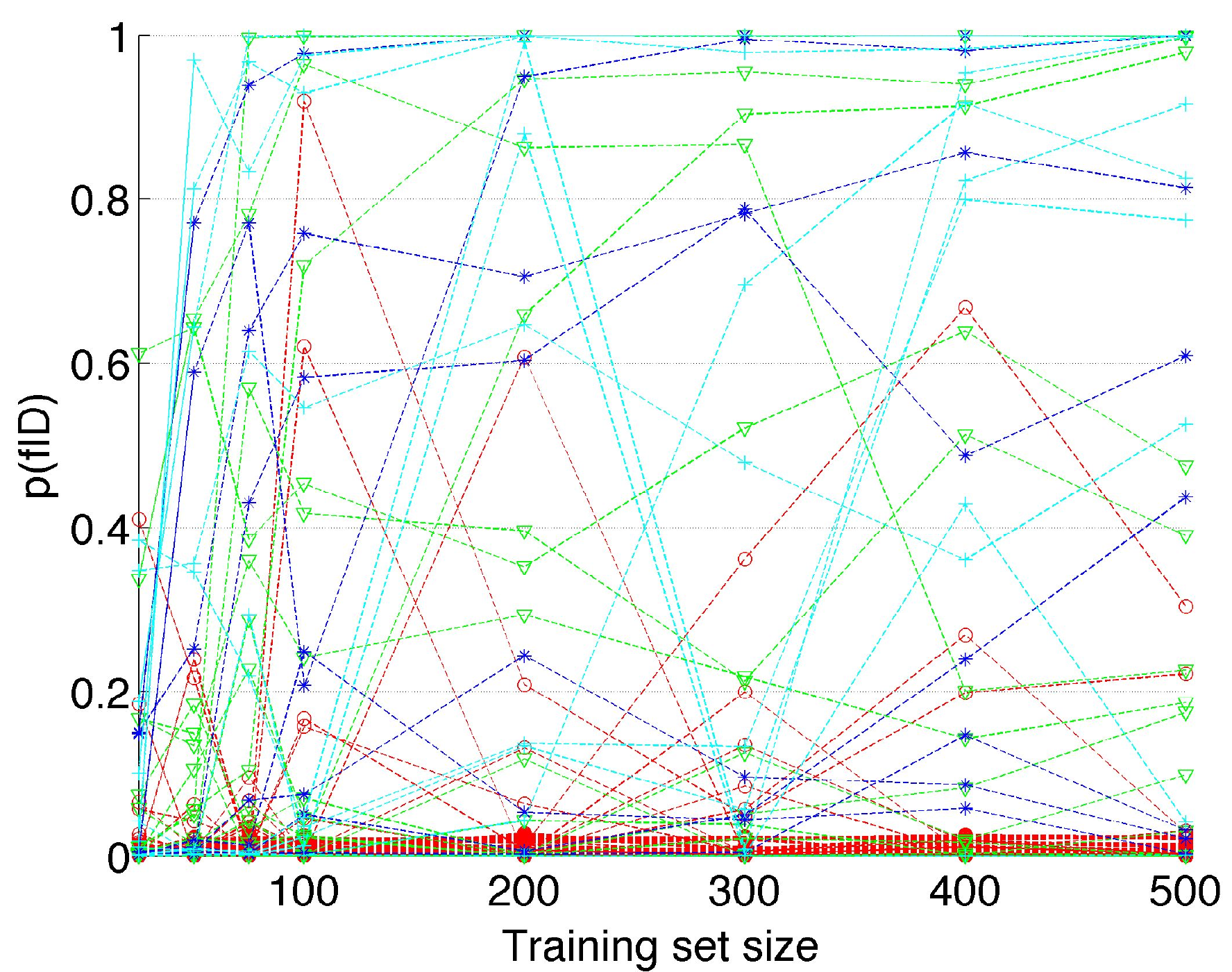}
	\label{fig:alarmPedgeMTL}}
	\hfil
\subfloat[POOL]
	{\includegraphics[width=0.32\textwidth]{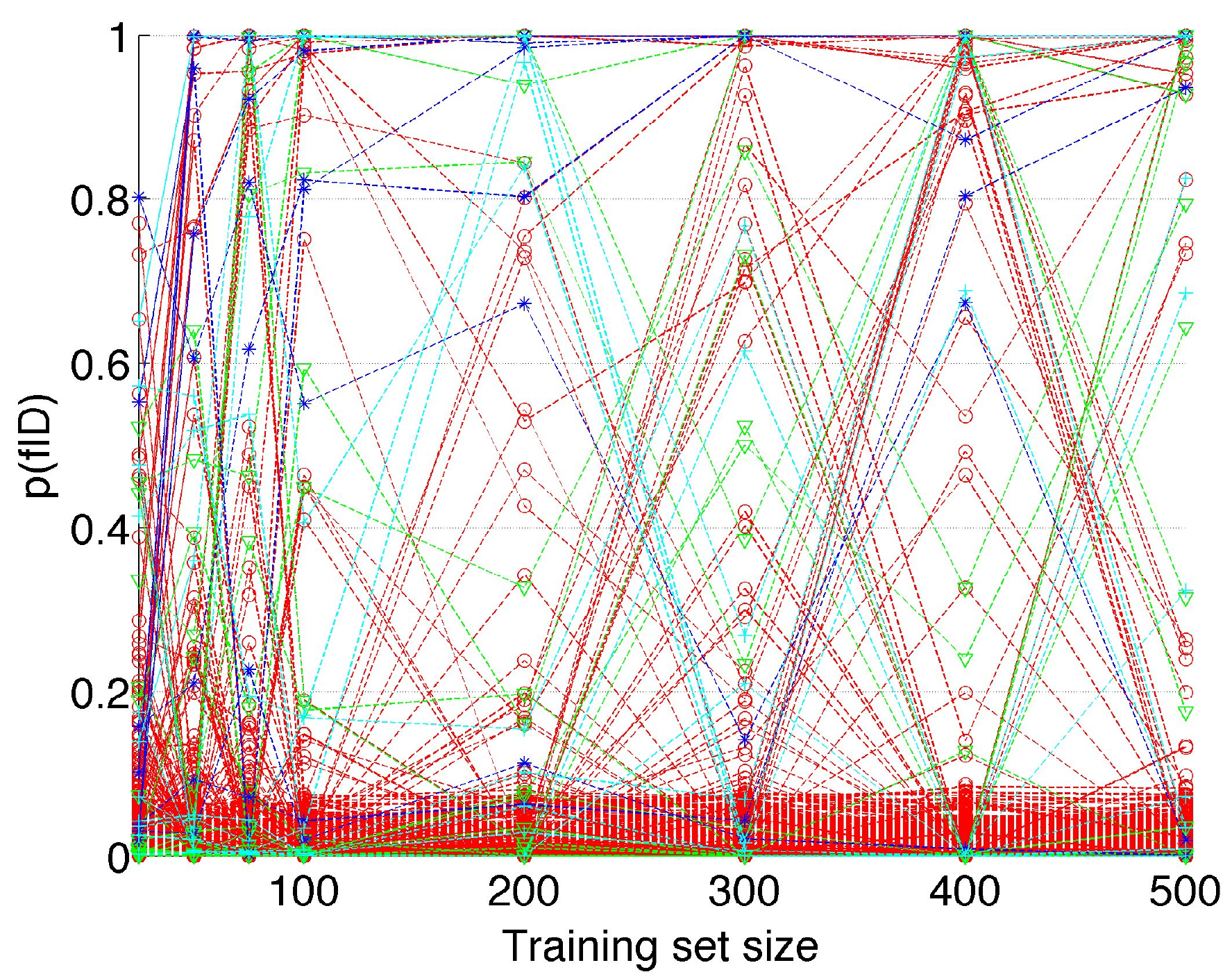}
	\label{fig:alarmPedgePOOL}}
\caption{Example posterior probability estimate for each edge in a modified \textit{alarm} network from various small sample sets (means calculated from 20 sample sets). Blue curves are true edges shared by both tasks, cyan curves are true edges unique to this task, green are reverse of a true edge, red are non-edges. Up is good for blue and cyan curves. Down is good for red and green curves.}
\label{fig:alarmPedge}
\end{figure*}

Area under the curve (AUC) summarizes the overall performance along the ROC curve. We report AUC for various amounts of training data in Table~\ref{table:auc} across 30 trials of the generative model. With these small training sets, the difficulty of the problem presented by each trial can vary quite a bit, therefore, the performance of each algorithm is compared directly on each trial by looking at how much greater the AUC is for MTL than the other algorithm. This increase in AUC score per trial is then averaged over all trials to give the numbers in Table~\ref{table:auc}. Furthermore, a paired-T test is performed to determine whether this increase in performance is significant at the 5\% confidence level. The winner of the paired-T test is given in Table~\ref{table:auc}.

Similar experiments are performed on the larger \textit{alarm} network. This network is too large for exact posterior computation, therefore we use MCMC approximation. We set MCMC hyper-parameters as recommended by \cite{Niinimaki11}, specifically, bucket size = 10, burn-in samples = 1000, sub-sample interval = 10 and total samples = 100. We tried other values (notably more samples, larger sub-sample interval, and longer burn-in) and found that they give nearly the same results. We also use the transfer approximation described in Section~\ref{sec:complexity}, with $h = 1000$. For the ground truth \textit{alarm} network, we use 10,000 training samples to estimate the true posterior of each feature, $P^*(f|D) = \hat{P}_{\mathit{STL}}(f|D_{10,000})$.

\begin{table}[tb]
\centering
\caption{Performance increase on \emph{alarm} for AUC given by MTL versus STL and MTL versus POOL}
\label{table:aucAlarm}
\begin{footnotesize}
\begin{tabular}{c || r | c || r | c}
  Training & \multicolumn{2}{c||}{MTL vs STL} & \multicolumn{2}{c}{MTL vs POOL} \\
  samples & \% incr & pair-t & \% incr & pair-t \\
  \hline
  25 & -0.12 & - & 0.04 & - \\
  50 & 0.29 & MTL & 1.76 & MTL \\
  75 & 0.43 & MTL & 3.63 & MTL \\
  100 & 0.67 & MTL & 4.49 & MTL \\
  200 & 0.05 & - & 5.19 & MTL \\
  300 & -0.07 & - & 8.68 & MTL \\
  400 &  -0.03 & - & 8.85 & MTL \\
  500 & -0.46 & STL & 10.45 & MTL \\
\end{tabular}
\end{footnotesize}
\end{table}

\begin{figure}[tb]
\centering
\subfloat[50 samples per task]
	{\includegraphics[width=0.48\columnwidth]{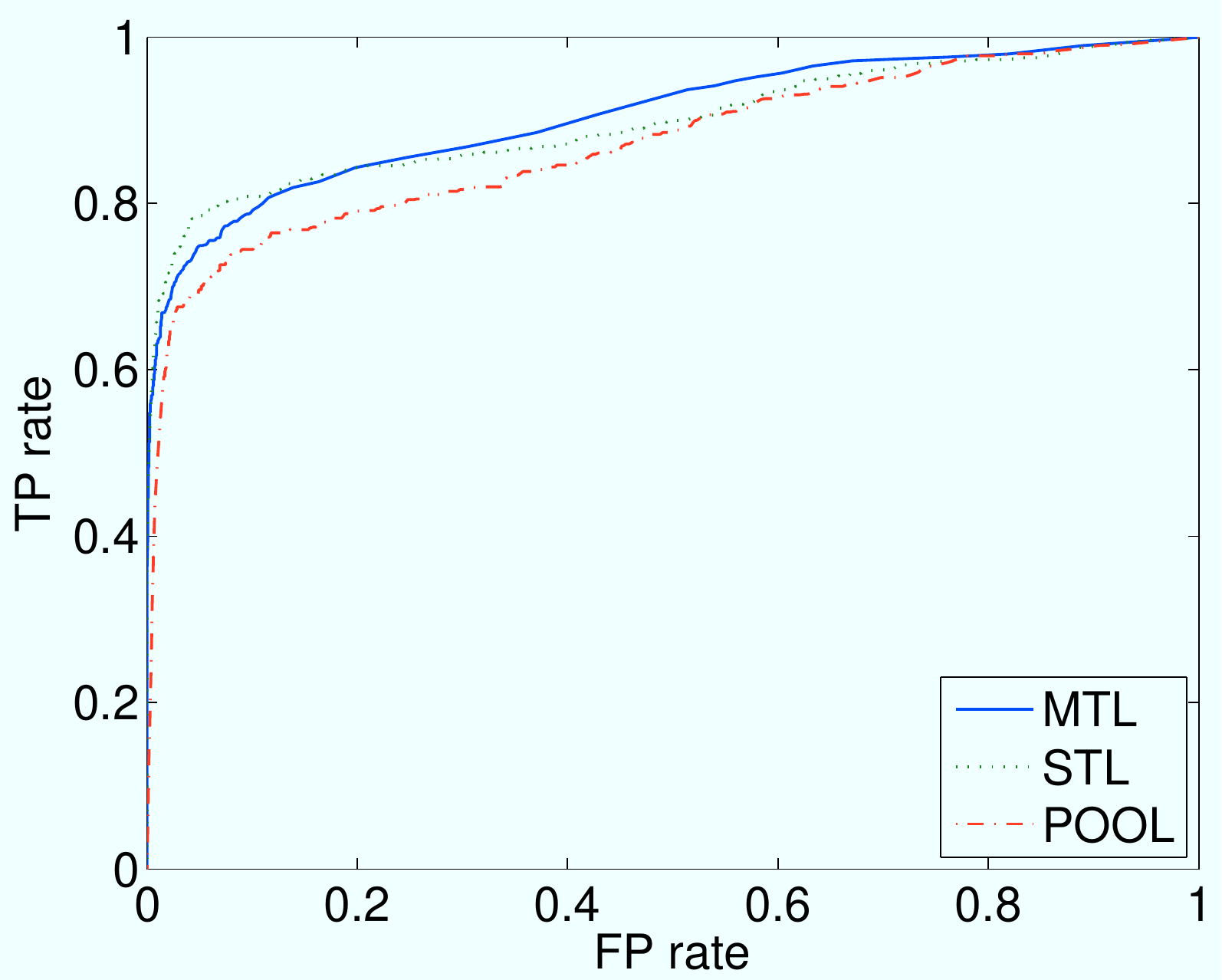}
	\label{fig:alarmRoc50}}
	\hfil
\subfloat[100 samples per task]
	{\includegraphics[width=0.48\columnwidth]{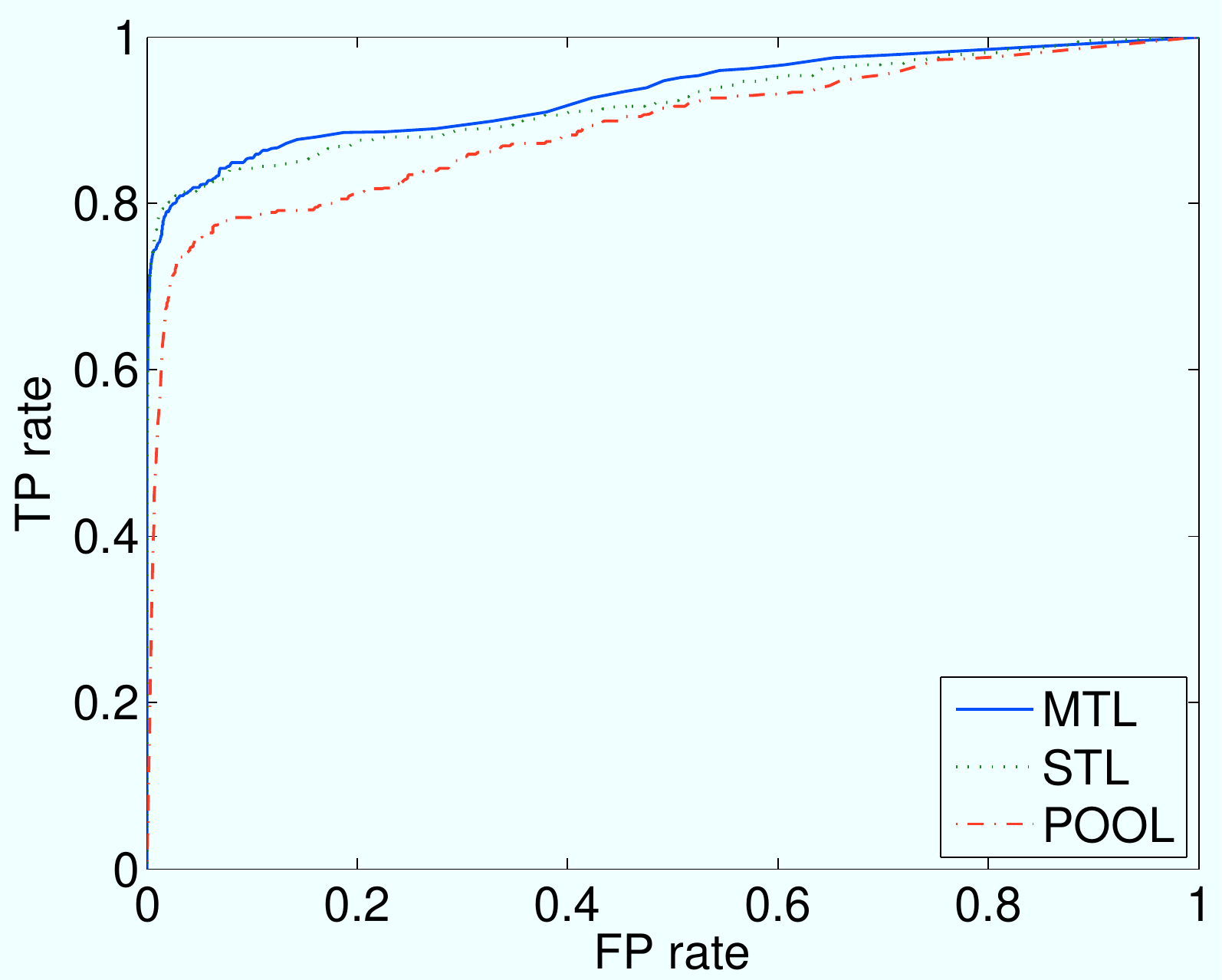}
	\label{fig:alarmRoc100}}
\caption{Example ROC curves for \textit{alarm} data. Each point represents the (FP rate, TP rate) aggregated over 2 tasks and 10 trials of the generative model, for a particular value of $\tau$.}
\label{fig:alarmRoc}
\end{figure}

The alarm network contains 37 variables, therefore there are 1,332 ordered pairs of nodes or potential edges in the set $E$. Of these, only 46 are true edges. We see again that on this data set, MTL estimates lower posteriors for the non-edges than STL or POOL (see Figure~\ref{fig:alarmPedge}). The ROC curves in Figure~\ref{fig:alarmRoc} show that POOL routinely identifies many false positives. The curves for MTL and STL are closer, but again MTL is better at reducing the number of false positives. This makes differentiating the true edges from the false edges easier at small training set sizes, see Table~\ref{table:aucAlarm} for AUC results. MTL dominates STL for small training sets. MTL dominates POOL at all training set sizes.






\section{Application to Neuroimaging}

Our goal is to find functional brain networks associated with schizophrenia. We start with functional magnetic resonance image (fMRI) data that measure the activity levels in regions of interest (ROI) in the brain. The activity level for each ROI is de-trended using a sliding window mean and then discretized into four levels representing Very Low, Low, High and Very High activity levels (relative to the mean activity level of that ROI). The functional brain network is modeled as a Bayesian network of information sharing using a multinomial of discretized activity level among ROIs. Data has been collected from 86 healthy control subjects (\emph{controls}) and 74 schizophrenia patients (\emph{patients}). For each subject, there are 384 full-brain scans which are the samples in our training data. Brain images are parcellated using the Talaraich atlas giving 150 ROIs. Therefore, for each subject, we have a $150 \times 384$ data matrix. We concatenate the data from several subjects to create the training data for each task. We apply both our MTL algorithm and the standard STL Bayesian structure discovery algorithm. As our goal is to identify different structures between tasks, we do not use the POOL method that learns identical structures for both tasks.

The number of subjects in this study is much larger than in many other studies that we are interested in. We would like to learn reliable networks from small studies, and so we sub-sample the subjects in this study to see how consistent our results are across subsets of subjects. For evaluation purposes, we use the full set of data (86 controls and 74 patients) to learn a large-sample model and use this learned model as the ground truth to measure the small-sample results against. The results show how well learned models over various subsets of subjects are representative of the larger \emph{control} and \emph{patient} populations.

We limit the size of the parent sets to $r = 2$. With this setting, the time to calculate family scores is approximately 3 hours. For MTL family score calculation, we use the approximation method described in Section~\ref{sec:complexity} with $h = $10,000. MCMC approximation is used to estimate the posterior likelihood of edges \cite{Niinimaki11}, with hyper-parameters bucket size = 10, burn-in samples = 5000, sub-sample interval = 10 and total samples = 1000.

\subsection{Small Samples}

\begin{figure}[tb]
\centering
\subfloat[Non-edges]
	{\includegraphics[width=0.48\columnwidth,bb=56 205 524 568]{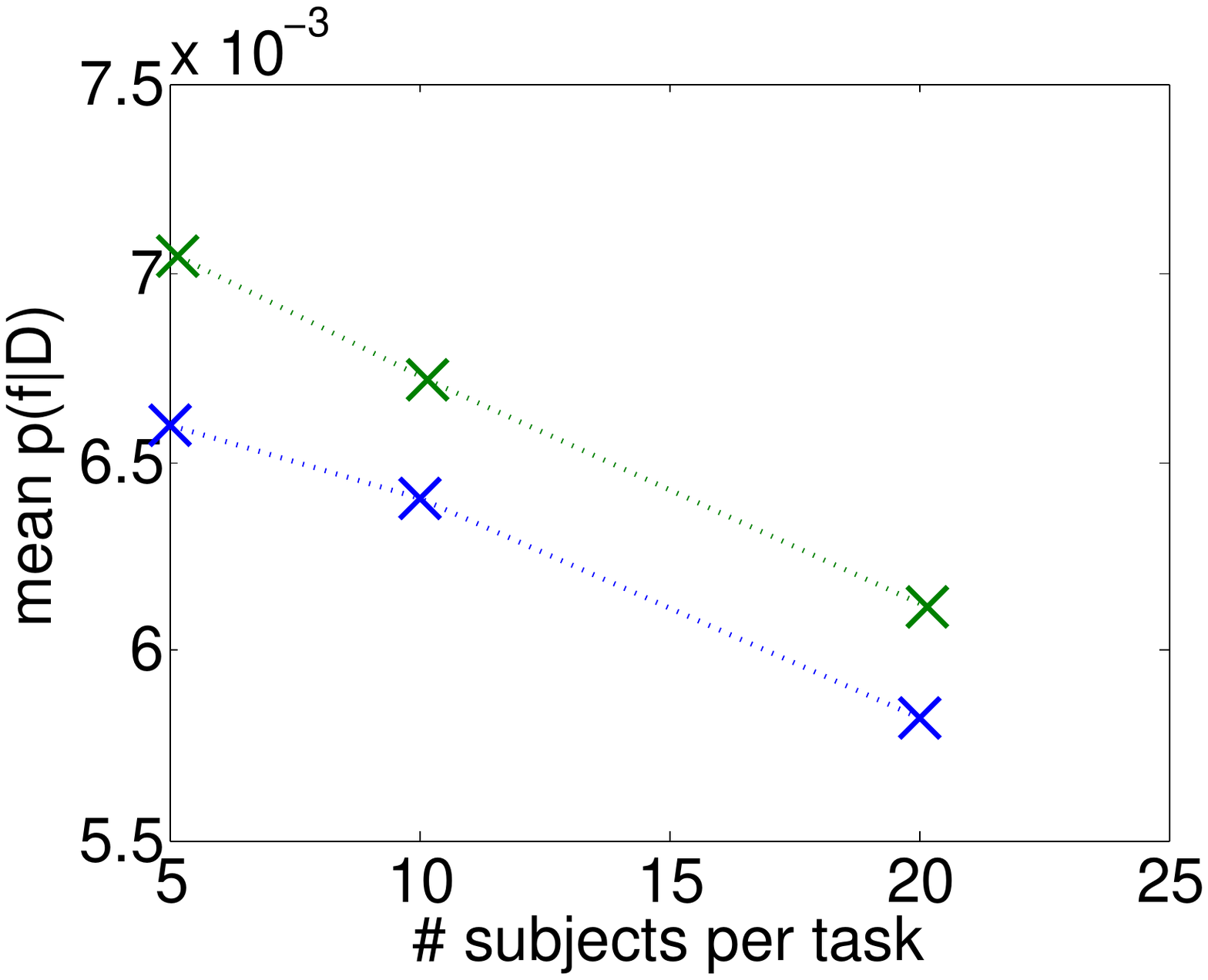}
	\label{fig:mcicNonedges}}
	\hfil
\subfloat[True edges]
	{\includegraphics[width=0.48\columnwidth,bb=56 205 524 568]{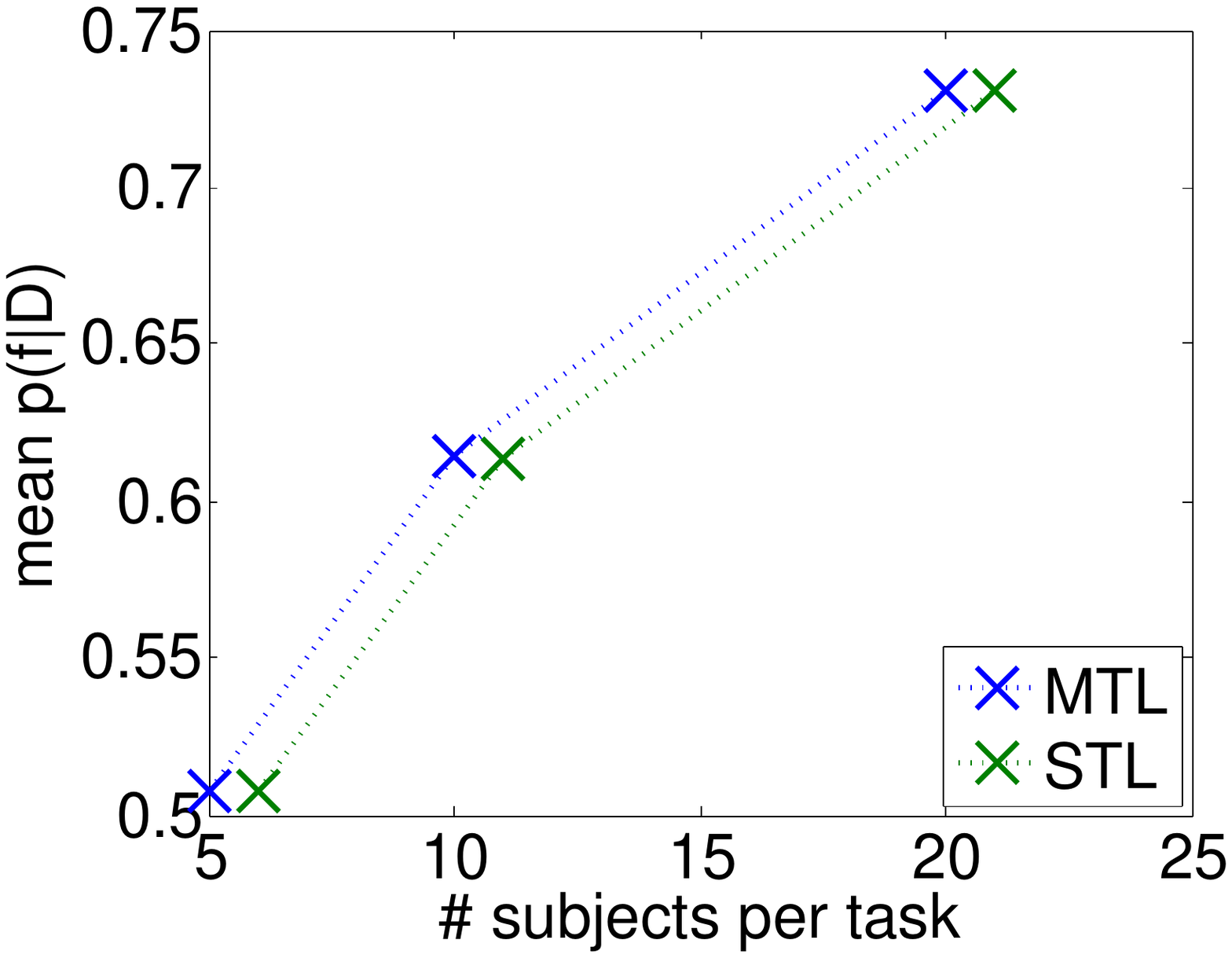}
	\label{fig:mcicTrueEdges}}
\caption{Estimated posterior of features from small subsets of subjects. Points are perturbed horizontally for visibility.}
\label{fig:mcicPedge}
\end{figure}

MTL estimates significantly lower posterior likelihoods on non-edges compared to STL. Evaluating results on real data is complicated by the fact that we do not have ground truth of known networks. Thereroe, we looked at the trend of estimates on smaller subset of the subjects and compare the results against the STL estimate from the full set of data. Figure~\ref{fig:mcicNonedges} shows that for edges that are \emph{not} determined to have real dependencies in the full data set (i.e. non-edges), the posterior estimate is significantly lower for MTL than STL. Significance was determined via a paired-t test at the 95\% confidence level over various subsets of subjects selected. Figure~\ref{fig:mcicTrueEdges} shows that there is no difference between MTL and STL in terms of the posterior estimate of true edges (according to paired-t test at 95\% confidence). Therefore, MTL is able to eliminate the non-edges with less data than STL, corroborating results from the benchmark (\emph{asia} and \emph{alarm}) networks.

\subsection{Learned Dependencies}

\begin{figure*}[tb]
\centering
\subfloat[Patient, 5 subjects]
	{\includegraphics[width=0.32\textwidth]{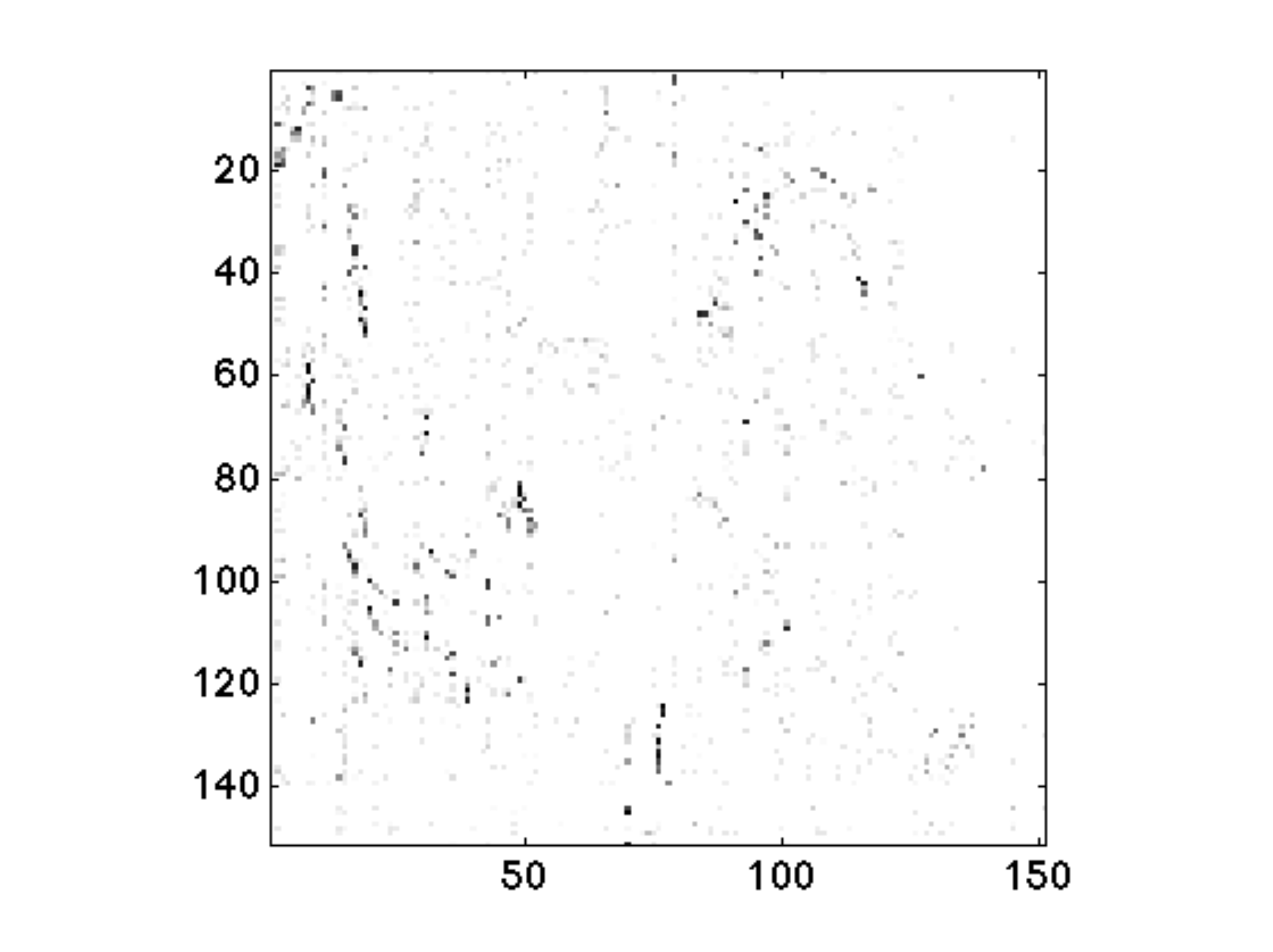}
	\label{fig:mcicPatient5}}
	\hfil
\subfloat[Patient, 20 subjects]
	{\includegraphics[width=0.32\textwidth]{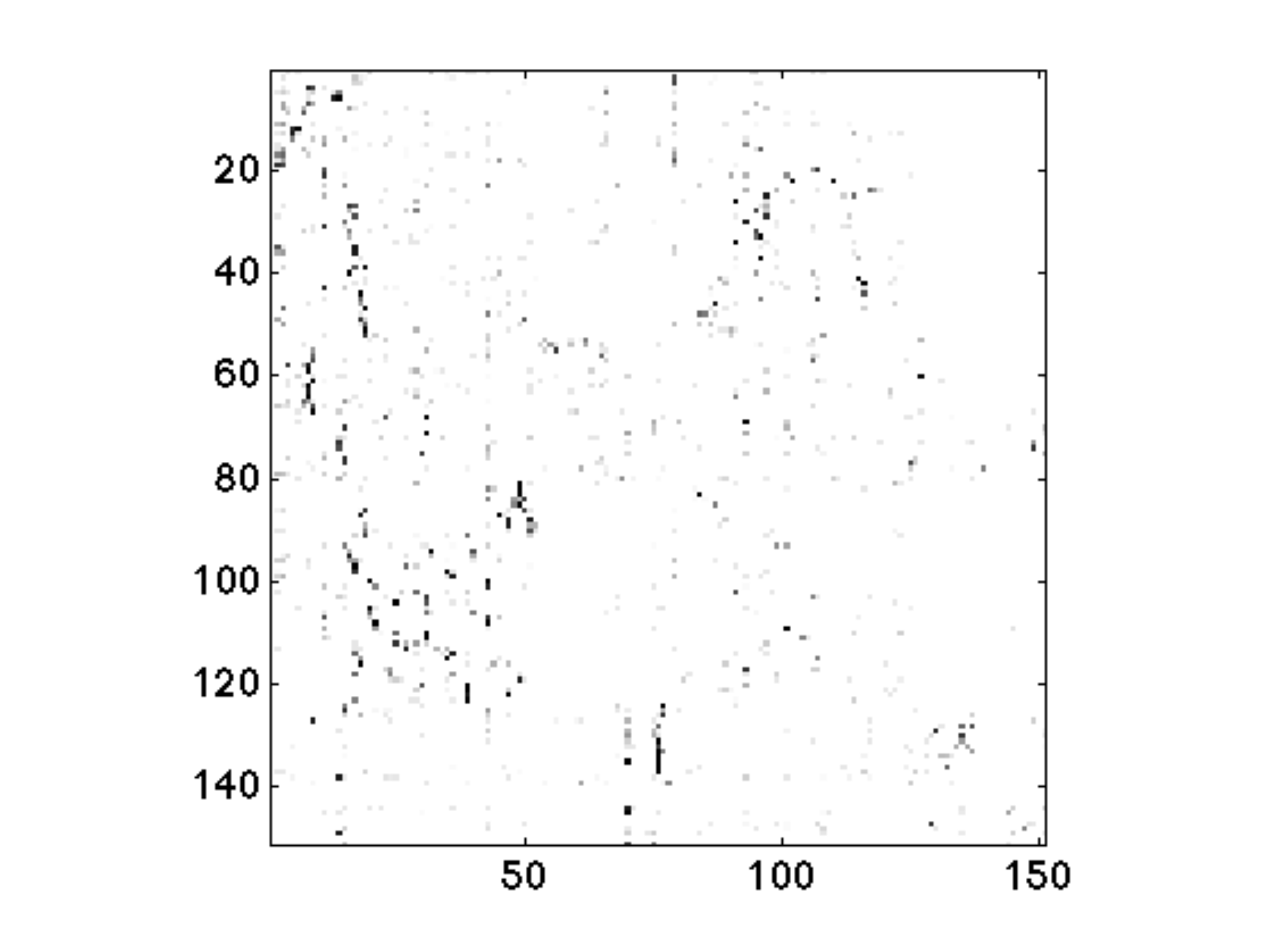}
	\label{fig:mcicPatient20}}
	\hfil
\subfloat[Patient, 74 subjects]
	{\includegraphics[width=0.32\textwidth]{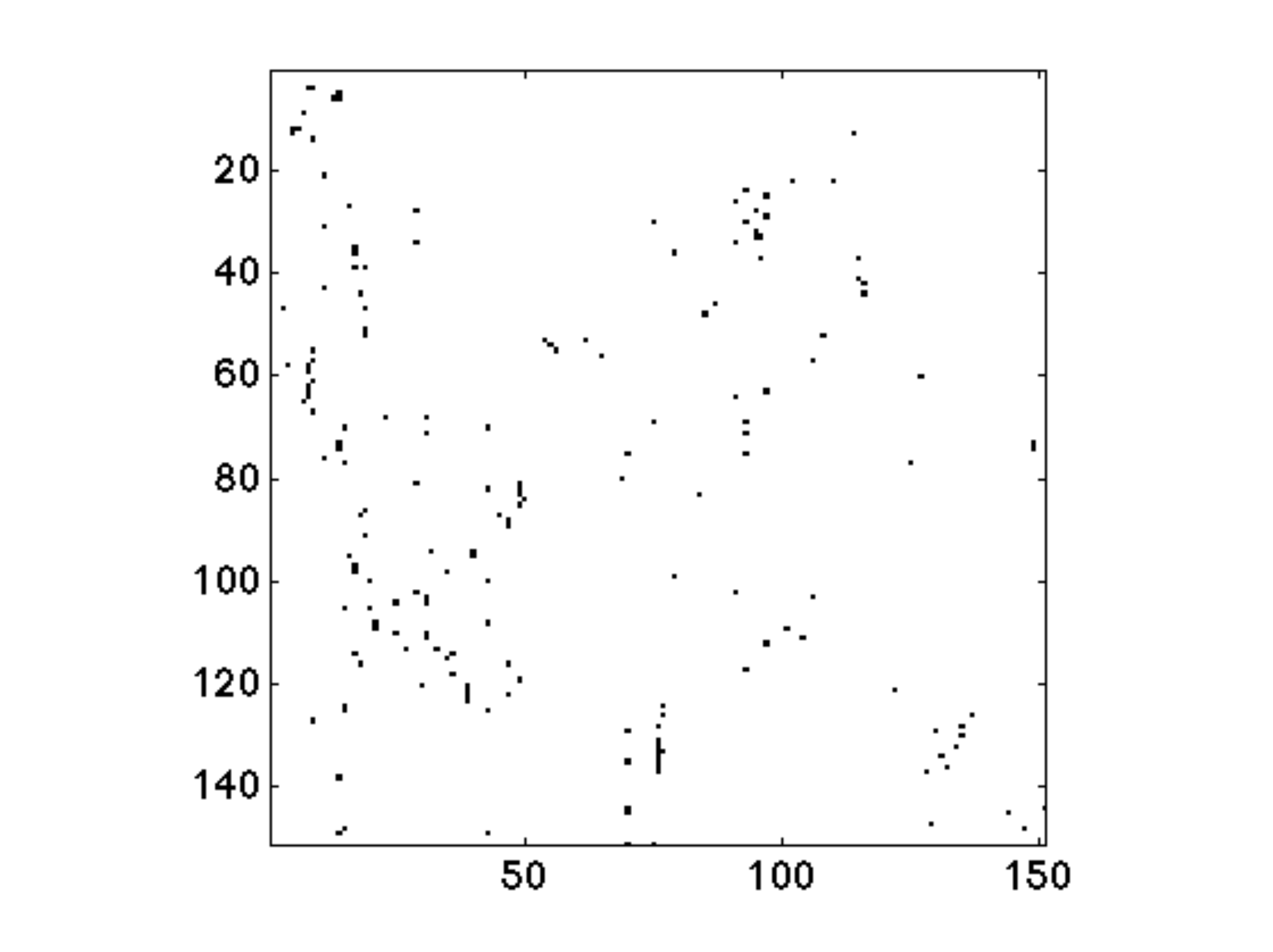}
	\label{fig:mcicPatientAll}}
	\\
\subfloat[Control, 5 subjects]
	{\includegraphics[width=0.32\textwidth]{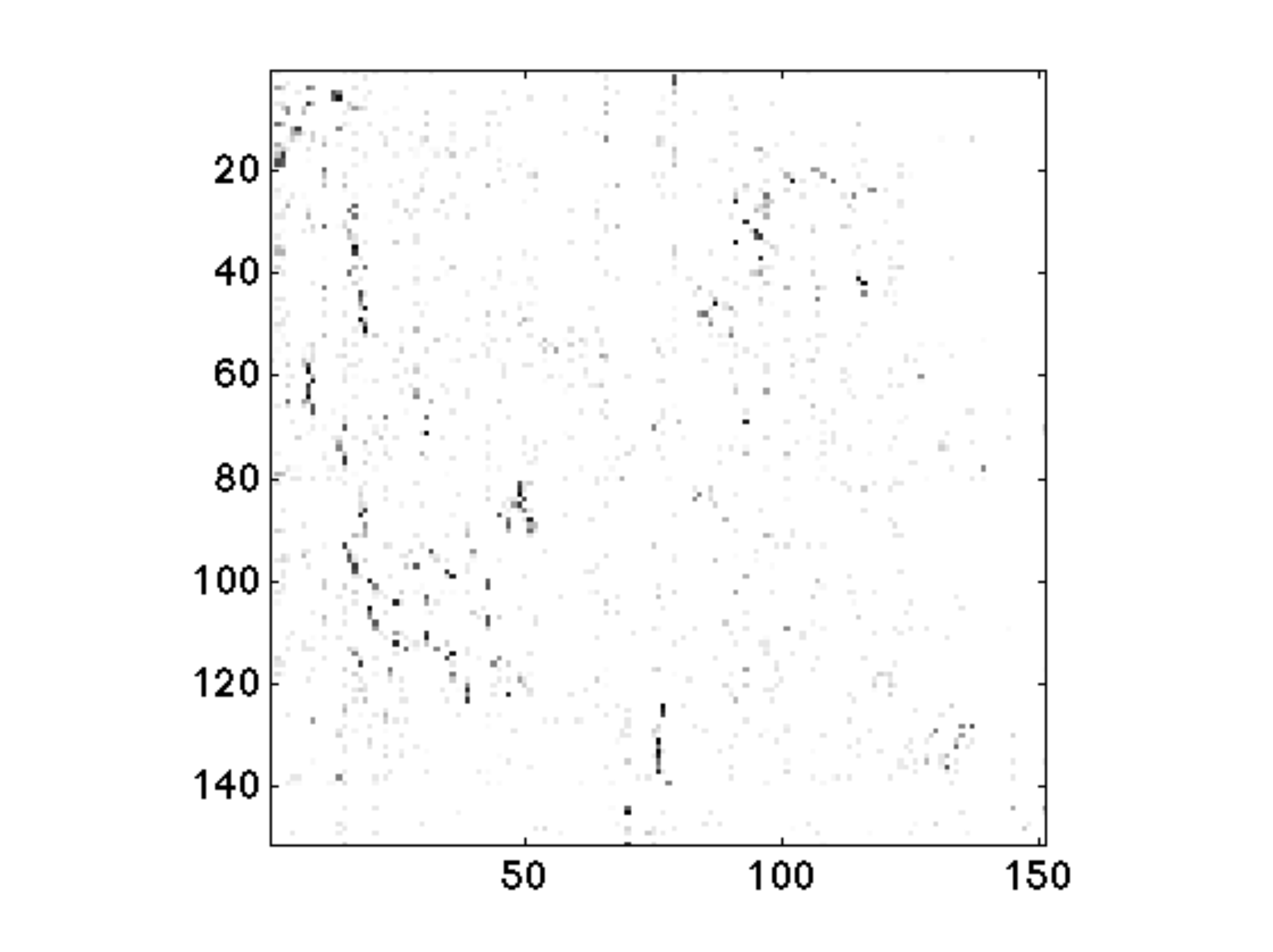}
	\label{fig:mcicControl5}}
	\hfil
\subfloat[Control, 20 subjects]
	{\includegraphics[width=0.32\textwidth]{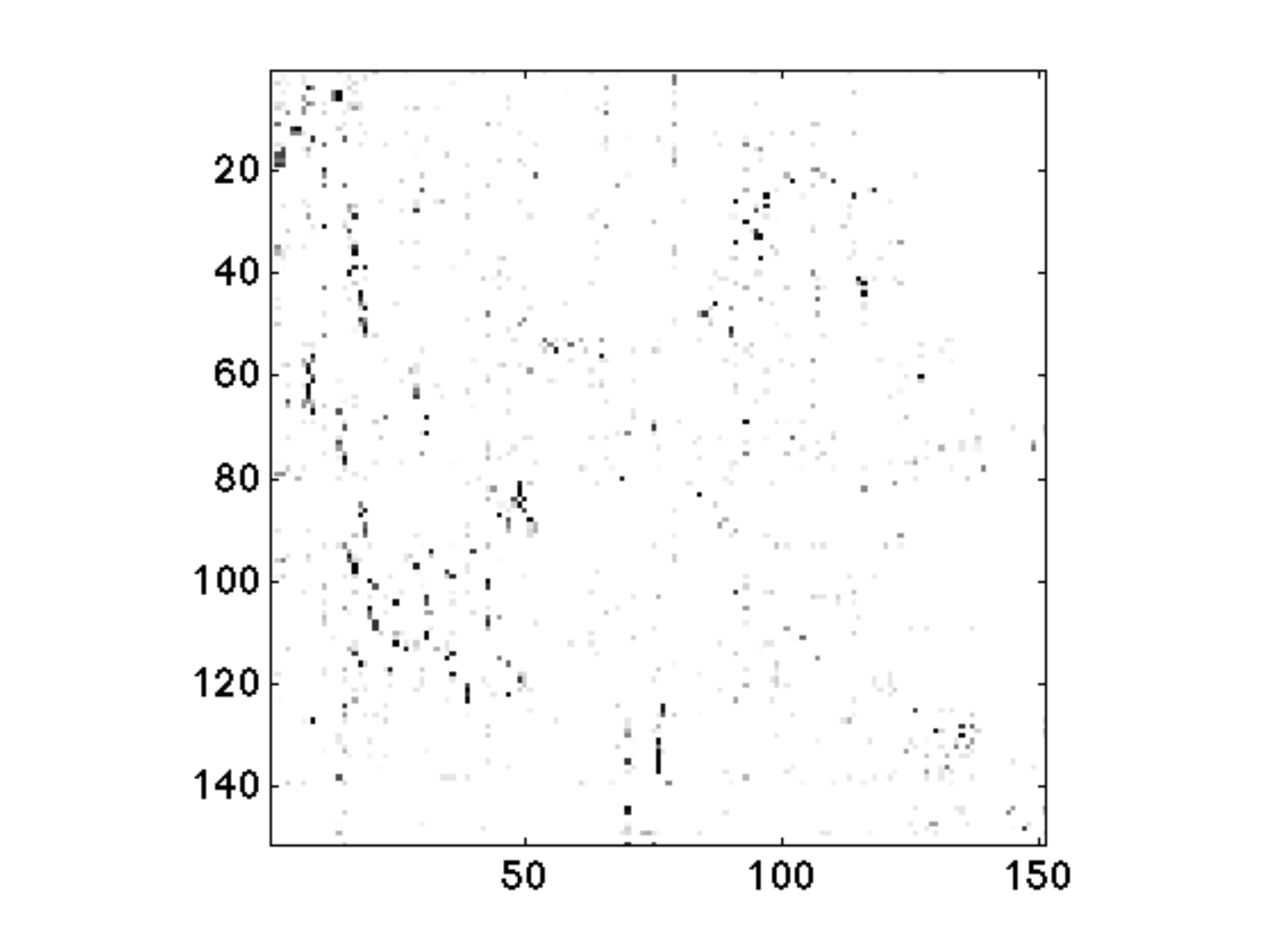}
	\label{fig:mcicControl20}}
	\hfil
\subfloat[Control, 86 subjects]
	{\includegraphics[width=0.32\textwidth]{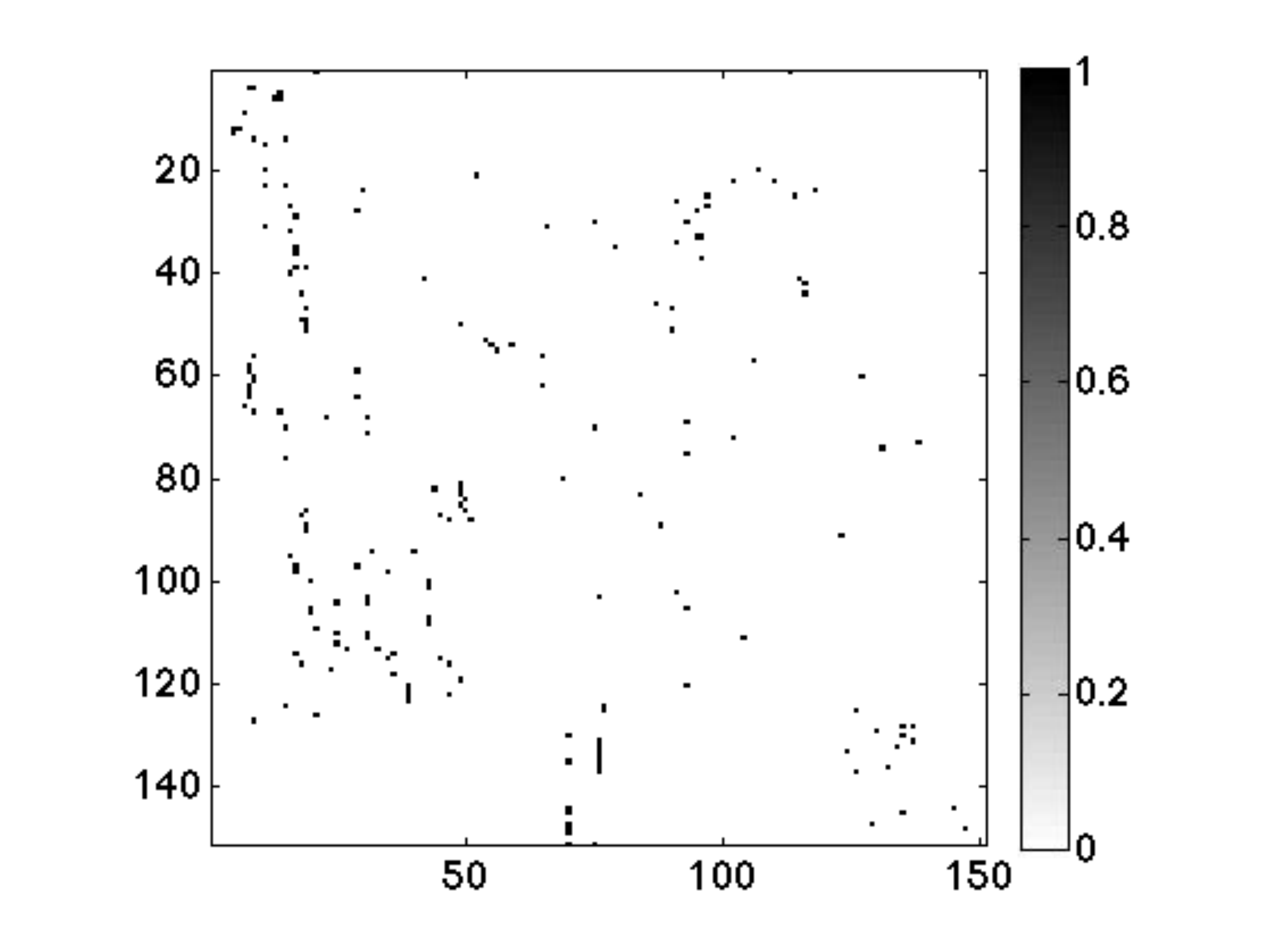}
	\label{fig:mcicControlAll}}
\caption{Posterior probability estimates for each edge learned from neuroimage data. Means calculated from 10 sample sets.}
\label{fig:mcicPedge}
\end{figure*}

In practice, rather than giving a complete network as a solution to the neuroscientists, the solution is presented as a list of likely dependencies or visualized using network layout software with edge thicknesses proportional to the probability of the dependency. The functional brain networks learned in this paper are large enough the static images are difficult to read. In this paper, we are more concerned with the robustness of learned models rather than the brain networks themselves; therefore, we display the edge likelihoods as an adjacency matrix. Figure~\ref{fig:mcicPedge} shows the mean of the posterior likelihoods across the 10 bootstrap samples of sets of subjects. High probability edges are black squares in the heatmaps. In this experiment, we see that overall the edge likelihoods learned are sparse and they are fairly consistent across subsets of subjects. The likelihoods become ``sharper" (closer to 0 or 1) as the amount of data increases, as expected. However, even for small numbers of subjects, we find clear patterns emerging. We also see that many dependencies are common to both the \emph{control} and \emph{patient} groups of subjects, while a few distinct differences are also visible. Through this type of visualization, a domain expert can gain insight into the possible interactions among variables in the system. The weight of the likelihood of each edge is important information to the domain scientist, which is not available from maximum a posterior multitask learning algorithms.

\section{Discussion}

Our structure bias in the order-modular framework for Bayesian network structure discovery can be applied to many other problems currently being researched. In this paper, we demonstrate the application of structural bias to the problem of multitask learning. We find promising results from our approach and expect that further improvements can be made by tailoring the bias term to the application. Additionally, more sophisticated methods for approximating the transfer bias on large networks could be explored.

Implementation of a structural bias in Bayesian structure discovery is critical for solving other problems as well. \cite{grzegorczyk_improving_2008} propose incorporating prior knowledge about biological networks in the form of a structural feature bias. Rather than using the exact calculation of posteriors that are possible when conditioning on orders, they attempt to find a different MCMC method for approximation. Their motivation was that it is inconvenient to define priors in the space of orders rather than structure. Our Theorem~\ref{thm:modularObjective} shows that it is indeed possible to define structural priors at the structure level to use the efficient algorithms that rely on conditioning on orders.

This structural bias term could also be used to transfer knowledge about the direction of Bayesian network edges from interventional experiments \cite{cooper_causal_1999}. Active learning of Bayesian network structure has been shown to significantly speed the learning of edges, particularly for getting directionality \cite{tong_active_2001-2}. Multitask active learning algorithms would be useful for transferring knowledge from an experiment where interventions are possible to a similar domain where such interventions may be more expensive or impossible. Recent work proposes principled methods for the transfer of causal relationships between domains \cite{Pearl2011Transportabilit}. Our paper provides a critical algorithmic mechanism to implement such transfer of knowledge.

\section{Conclusions}

We have presented a multitask Bayesian network structure algorithm. This algorithm is able to successfully leverage data from related tasks to improve the estimate of network structure features given limited amounts of data. The primary contribution is determining that structural priors that are order-modular can be used to impose inductive bias among tasks. By using local structural priors, we achieve three goals simultaneously: 1) an intuitive inductive bias at the level of structures rather than orders;  2) take advantage of the most efficient structure discovery algorithms; and 3) closed form Bayesian model averaging over the transfer strength parameter. Empirical evidence suggests that multitask learning of Bayesian networks reduces the number of spurious dependencies learned, particularly at small sample set sizes.

\subsubsection*{Acknowledgments}

Thanks to Vincent Clark and the Mind Research Network for providing data and interesting data mining problems. Also, thanks to Eric Eaton and Paul Ruvolo for helpful discussions. Work funded by a grant from ONR N000141110139.

\bibliographystyle{IEEEtran}
\bibliography{MyLibrary}

\end{document}